\pdfoutput=1
\documentclass[11pt]{article}

\usepackage[preprint]{acl}

\usepackage{times}
\usepackage{latexsym}

\usepackage[T1]{fontenc}
\usepackage[utf8]{inputenc}

\usepackage{microtype}

\usepackage{inconsolata}

\usepackage{graphicx}

\usepackage{array}
\usepackage{booktabs}
\usepackage{hyperref}
\usepackage{float}

\title{Investigating Cost-Efficiency of LLM-Generated Training Data \\for Conversational Semantic Frame Analysis}

\author{
  Shiho Matta$^{\dagger}$, 
  Yin Jou Huang$^{\dagger}$, 
  Fei Cheng$^{\dagger}$, 
  Hirokazu Kiyomaru$^{*}$, 
  Yugo Murawaki$^{\dagger}$ \\
  $^{\dagger}$Kyoto University \\
  $^{*}$National Institute of Informatics \\
  \texttt{\{matta, huang\}@nlp.ist.i.kyoto-u.ac.jp}, \texttt{\{feicheng, murawaki\}@i.kyoto-u.ac.jp}, \\
  \texttt{kiyomaru@nii.ac.jp}
}

\begin{document}
\maketitle
\begin{abstract}
Recent studies have demonstrated that few-shot learning allows LLMs to generate training data for supervised models at a low cost. However, the quality of LLM-generated data may not entirely match that of human-labeled data. This raises a crucial question: how should one balance the trade-off between the higher quality but more expensive human data and the lower quality yet substantially cheaper LLM-generated data? In this paper, we synthesized training data for conversational semantic frame analysis using GPT-4 and examined how to allocate budgets optimally to achieve the best performance. Our experiments, conducted across various budget levels, reveal that optimal cost-efficiency is achieved by combining both human and LLM-generated data across a wide range of budget levels. Notably, as the budget decreases, a higher proportion of LLM-generated data becomes more preferable.
% \begin{itemize}
%     \item LLM-generated data is rapidly developing
%     \item challenge: LLM-generated data has relatively low accuracy 
%     \item research question: how to address the trade-off between human data and LLM-generated data?
%     \item contribution: LLM-generated data, demonstrated that the optimal cost-efficiency appears when combining human data and LLM-generated data
% \end{itemize}

% [need rewriting] Collecting annotated data to fine-tune a in-house model can be costly. It has been shown that using LLM-generated training data can replace such human-labeled data in some tasks. However, there lacks a general observation of how much LLM-generated data you should incorporate given a limited amount of budget, especially when LLM's raw labeling ability is lower than that of humans. In this paper, we investigate the cost-efficiency of LLM-generated training data in a fixed budget setting, finding the optimal solution to borrow the strength of LLM based on your budget.
\end{abstract}

\section{Introduction}
% ***DEPRECATED***
% \begin{itemize}
%     \item Including LLM-generated data in your training data for a smaller supervised model on a downstream task is a common practice and for simple tasks it is shown effective.
%     \item However, LLMs don't perform so well on IE tasks. Therefore the accuracy of generated data also has limited coherence to human labels.
%     \item It is obvious that when you have plenty of money, you should collect human-labeled data for best performance. But still, we don't know how LLM data perform when you have limited data and want to bootstrap performance of a new task or into a new domain. 
%     \item In this paper, we investigate the effectiveness of LLM-generated data in SFA. (definition of SFA). For this task, the preliminary experiment shows that only the largest GPT-4 model (GPT-3.5, GPT-4-Turbo all fail to follow instructions) yields usable results.
%     \item We show that: 1. it is useful to include LLM data when tackling an IE task with a limited budget. 2. an indicator of when you should rely on LLM data, combine LLM data and human data, and stop collecting LLM data. 3. If you don't need human text data, you can generate the text using an LLM and still get reasonable performance, as label accuracy is the main bottleneck in this case.
% \end{itemize}

% 1. data is crucial for SLM model
% Creating a supervised learning model (SLM) requires high-quality data. However, the construction of such data with human annotation is costly.
It is costly to construct training data with human annotation for supervised learning models (SLMs). In recent years, large language models (LLMs) like GPT-4 have demonstrated remarkable abilities in generating coherent text, understanding context, and following complex specifications to accomplish tasks \cite{brown2020language, openai2024gpt4technicalreport}. Therefore, there have been many attempts to leverage existing LLMs as data annotators to generate training data for SLMs, aiming to reduce data costs. Studies have indicated that using LLM-generated data can cut costs significantly while maintaining a reasonable performance against human-annotated data for certain tasks \cite{wang-etal-2021-want-reduce, ding2023gpt3}. 

\begin{figure}[tbp]
  \includegraphics[width=\linewidth]{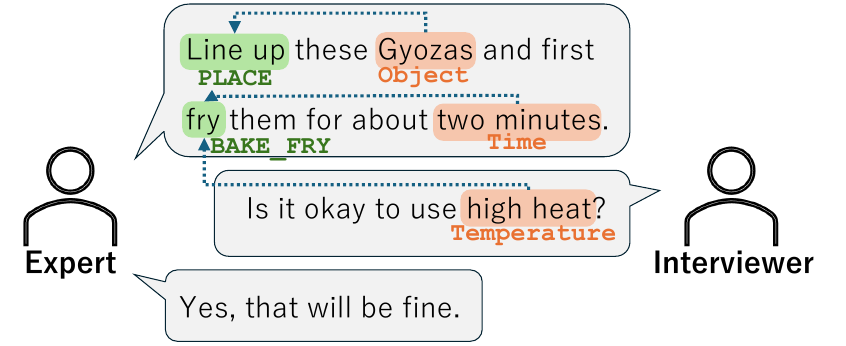} 
  \caption {A dialogue piece with semantic frame annotation. Green indicates a trigger, and orange indicates an argument. The argument-trigger relation is illustrated with arrows. This is a simplified demonstration translated from Japanese. }
  \label{fig:dialogue_example}
  \vspace{-5mm}
\end{figure}

% SFA is an example. 
In this paper, we focus on the task of analyzing semantic frames in Japanese technical expert-interviewer dialogues in the EIDC dataset \cite{okahisa_2022_constructing,chika_2024_domain}. Semantic frame analysis (SFA) captures salient knowledge exchanged between speakers by extracting semantic frames, which represent events within a given context.
% , which is essential for understanding the technical details in the dialogue. Specifically, SFA aims to extract semantic frames that represent specific events mentioned in the context. 
A semantic frame consists of a \textbf{trigger}, which is a predicate that represents the main action of the event, and \textbf{arguments} of the trigger, which are the details of the event. In Figure~\ref{fig:dialogue_example}, two semantic frames are annotated: "line up" (frame type PLACE) and "fry" (BAKE\_FRY). The first frame has one Object argument, while the second has Time and Temperature as arguments. Colloquial interview dialogues often contain repetitions and confirmations of technical details, and as shown, the interviewer's question introduces a new argument to the frame.
% To be noted, there are many repetitions and confirmations of the technical details due to the highly colloquial nature of technical interview dialogues. Usually, the interviewer asks questions to better understand details, which is the case in Figure \ref{fig:dialogue_example} where the interviewer's question adds a new argument to the frame. 
% Therefore, the mainstream approach for SFA is carried out on supervised models that can capture entity token positions to process details scattered across sentences \cite{chika_2024_domain, ebner-etal-2020-multi}. 

\begin{figure*}[t!]
\includegraphics[width=\linewidth]{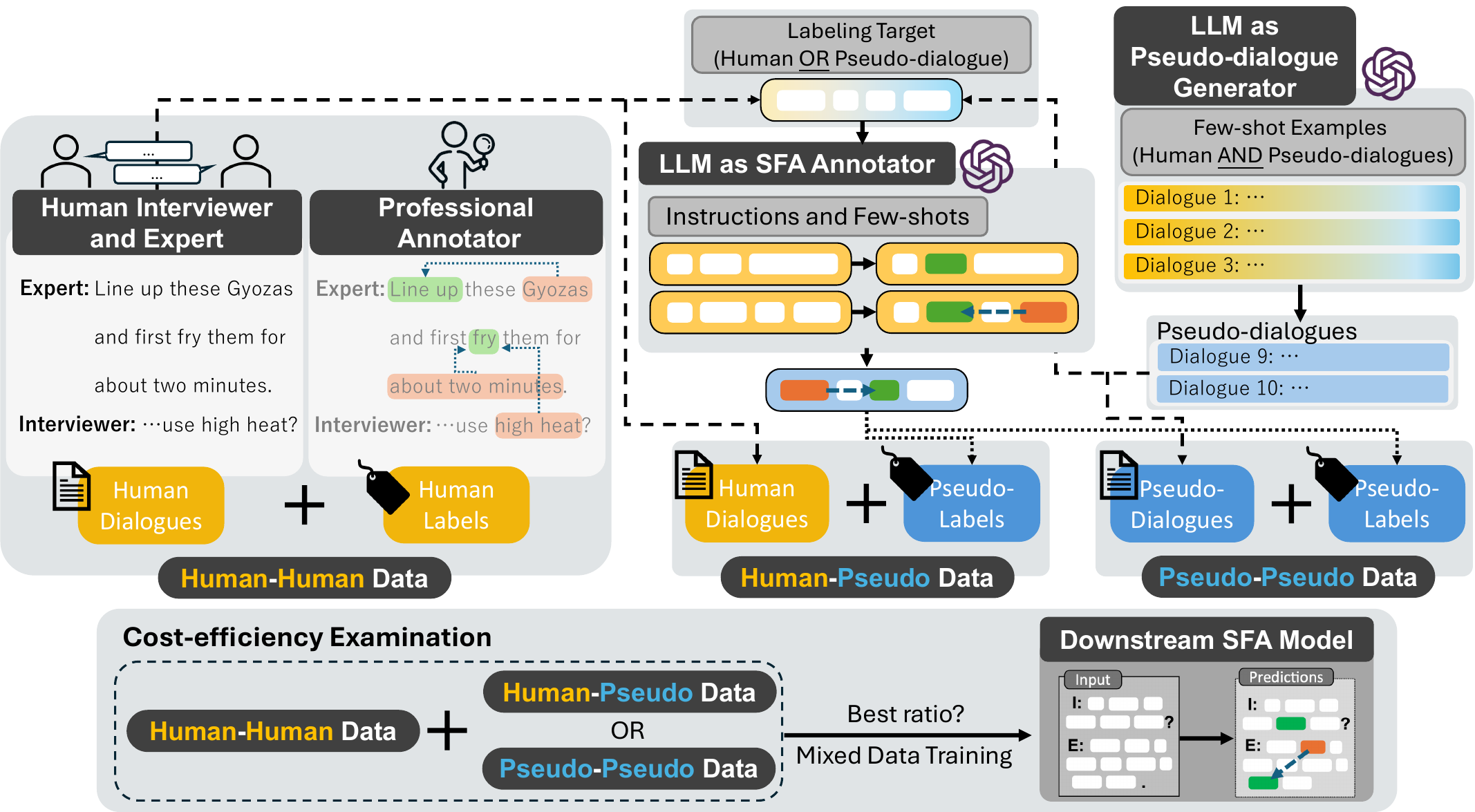}
  \caption {The overview of our proposal to create two types of LLM-generated data: Human-Pseudo and Pseudo-Pseudo, and to investigate the cost-efficiency of combining them with human-labeled data under different budgets. The dialogue example is translated from Japanese. }
  \label{fig:overview}
  \vspace{-5mm}
\end{figure*}

Human-annotated data is typically expensive, and the EIDC dataset is no exception. The collection of one dialogue and its semantic frame annotation cost approximately \$133 (Chika, personal communication, 01/2024). % For individual researchers, doctoral students, and small enterprises with limited budgets, it would be challenging to collect a sufficient amount of training data. As one example, 
On the other hand, the average annual research grant for doctoral students at Japanese universities is approximately \$4,000. Even if the entire amount were allocated to data collection, it would only yield 30 dialogues, which should not be optimal for supervised learning. 
In contrast, a GPT-4-generated dialogue and SFA label pair in our experiments cost only \$3. 

Although LLM-generated data is cheap, it typically has lower labeling accuracy than human-labeled data in certain tasks such as named entity recognition and relation extraction, which are similar to SFA \cite{wang-etal-2021-want-reduce, ma-etal-2023-large}. This raises the question: with a limited budget, can allocating a portion of it to LLM-generated data improve performance on SFA? 
% While there are numerous attempts to utilize LLMs to generate training data, \citet{ma-etal-2023-large} noticed that few-shot LLMs generally underperform on specific tasks such as event detection and event argument detection, which are simpler forms of trigger detection and argument detection in SFA, as they only extract relation triplets in their setup. 
% We expect that in SFA, the accuracy of human data will exceed LLM-generated data when the budget is sufficient. 
We answer this by training an SLM for SFA with a combination of more accurate human data and cheaper LLM-generated data. 
% In other words, we assume that with a given budget, there is an optimal ratio to combine human and LLM-generated data that yields the best performance on SFA. 
We set budgets ranging from as low as \$200 to up to \$12,800, which is the 3-year sum of the aforementioned average annual research grant for doctoral students at Japanese universities. For each specific budget, we set different ratios of human data and LLM-generated data to train the SFA model to search for optimal cost-efficiency. % Compared to previous studies that separately compared human data with LLM-generated data@cite, we believe our setting is more reflective of real-world scenarios.

% 4. LLM as pd generator and SFA labeler (not too detailed). we used few-shots, human-labeled data as seeds for both pd and SFA. we set different budgets and ratios. there is no current work that focus on sequence labeling. 

% 1. data: dialogue + label
% 2. an LLM can be used for both dialogue generation and label synthsize
% 3. Thus, we apply pseudo-label to existing human dialogue and get Human-Pseudo
% 4. If we generate pseudo-dialogue and apply pseudo-label both using an LLM, we get Pseudo-Pseudo data.

We create two types of LLM-generated data: Human-Pseudo and Pseudo-Pseudo, as illustrated in the overview Figure~\ref{fig:overview}. Human-Pseudo is comprised of human dialogues and pseudo-labels applied by GPT-4, and Pseudo-Pseudo contains both pseudo-dialogues and pseudo-labels. To construct pseudo-dialogues, we follow the self-instruct strategy \citep{wang_2023_selfinstruct} to generate new and diverse dialogues starting from a few reserved human dialogues. We also utilize GPT-4 as the SFA labeler by providing few-shot labeling examples. 
%Notably, to our knowledge, this is the first approach to design a novel prompting scheme that allows an LLM to explicitly label the spans of different types of entities simultaneously in a sequence labeling task like SFA. 
Notably, we propose a novel prompting scheme that enables an LLM to handle SFA by (1) explicitly managing entity positions to capture entities scattered across multiple utterances, and (2) facilitating the conversion of output data into a sequence-labeling SLM.

% utilize an LLM to generate sequence-labeling style data for SLMs in an end-to-end manner, which we think is essential for tasks like SFA. 

% 5. contribution
Our empirical results indicate that, across a range of budgets, incorporating LLM-generated data into the training data helps reach optimal cost-efficiency. The lower the budget is, the more LLM-generated data should be included for best performance. 
% Furthermore, our findings suggest that at budget extremes—either significantly high or low—reliance should be placed entirely on either human-annotated data or LLM-generated data, respectively. 
Another key contribution of our work is the direct comparison between LLM-generated data with human text and LLM-generated text (Human-Pseudo vs. Pseudo-Pseudo).
% evaluation of the effectiveness of using LLMs to generate both text and labels for SFA. 
We demonstrate that, even for a task requiring text data like technical interviews, LLM-generated text can be used without significantly compromising downstream task performance. 

\section{Related Work}

% sfa and sfa-like tasks
% method: cascaded sequence labeling
% recently: joint sequence labeling model
% remove subtask
\textbf{Semantic Frame Analysis (SFA) in Dialogues.} 
Semantic frame analysis is a task inspired by frame-semantic parsing (FSP) and semantic role labeling (SRL). Unlike the FrameNet project used in FSP \cite{baker-etal-1998-berkeley-framenet} or PropBank used in SRL \cite{kingsbury-palmer-2002-treebank}, the frame design in semantic frame analysis differs in two key ways: (1) the trigger type is curated for each topic domain and is predicate-centered, and (2) the argument types are common among different domains. 
% In previous studies on FSP and SRL, various methods—usually cascaded methods are used to identify triggers (a.k.a targets) and arguments in multiple steps \cite{johansson_2008_dependencybased,das-etal-2010-probabilistic,swayamdipta2017framesemanticparsingsoftmaxmarginsegmental, kalyanpur2020opendomainframesemanticparsing}. 
Here, we refer to the process of identifying the span and type of triggers and arguments as \textbf{Trigger Detection} and \textbf{Argument Detection}.  

Frame semantics can be used to capture critical information in dialogue situations. \citet{skachkova-kruijff-korbayova-2021-automatic} proposed using frame semantics in the domain of disaster response. The extracted information is used to capture and interpret verbal team communication for mission process assistance. In this work, we focus on conversational SFA in Japanese interview dialogues, specifically the cooking section of the EIDC dataset \citep{okahisa_2022_constructing, chika_2024_domain}.

% Previously, tasks like FSP and SRL were solved using probabilistic models \cite{das-etal-2010-probabilistic} and RNN-based models \cite{swayamdipta2017framesemanticparsingsoftmaxmarginsegmental}.
% Both \citet{das-etal-2010-probabilistic} and \citet{swayamdipta2017framesemanticparsingsoftmaxmarginsegmental} adopted frame designs that are derived from FrameNet \cite{baker-etal-1998-berkeley-framenet}. 
\citet{ebner-etal-2020-multi} tackled argument detection in a multi-sentence setting to better capture events that span across sentences, which is similar to our setting that is done on the dialogue level. \citet{kalyanpur2020opendomainframesemanticparsing} introduced Transformer-based \cite{vaswani2023attentionneed} models to FSP. They used a seq-to-seq Transformer model and formulated FSP as a text generation task by tagging entities using token index numbers, specifically for arguments. 
% \citet{wang-etal-2022-mrc} solves argument detection using the machine reading comprehension (MRC) approach with a sequence-labeling model. 

In this study, we adopt JaMIE \cite{cheng-etal-2022-jamie} as our SFA SLM. With its sequence-labeling nature and a relation decoder, it can solve trigger and argument detection at one time, making it an end-to-end solution for SFA.

% no existing work on SFA
% no existing work focus on a budget-wise cost-efficiency setting 
\textbf{LLMs for SFA-like tasks.} While no existing work directly targets SFA using LLMs, recent studies have explored related tasks, such as named entity recognition (NER) and relation extraction (RE).
% Previous studies have employed LLMs to tackle similar tasks like semantic parsing, which extracts components from a given context and forms them into a task-specific representation for QA tasks \citep{mekala-etal-2023-zerotop} or SQL queries \citep{drozdov2022compositionalsemanticparsinglarge}. 
% \citet{cheng2024potentiallimitationsllmscapturing} discussed the potential and limitations of LLMs on SRL. 
% Recent studies have also investigated the use of LLMs for similar tasks such as named entity recognition (NER), relation extraction (RE), event extraction (EE), etc. 
\citet{wang2023gptner} reformulated NER as a text-generation task by wrapping entities in tag pairs, allowing LLMs to process them efficiently.  
\citet{zhang2023aligning} and \citet{wan2023gptre} enhanced LLM performance on RE tasks by improving prompt design. 
\citet{sun2023pushinglimitschatgptnlp} tackled various NLP tasks, including NER and RE, by utilizing improved prompting and few-shot retrieval methods, similar to the approaches in \citet{wang2023gptner} and \citet{wan2023gptre}. 
These studies, along with the method proposed by \citet{kalyanpur2020opendomainframesemanticparsing}, have inspired our prompt design for SFA using an LLM (Figure~\ref{fig:sfa}). 
% Our design not only facilitates efficient LLM processing but also allows for the straightforward conversion of generated outputs into training data for our in-house model, as the output closely resembles the sequence labeling format. \citet{agrawal2022large_clinical_ie_llm} demonstrated the feasibility of using few-shot GPT-3 to extract clinical information from medical notes into a structured text pattern. 

Meanwhile, many studies have also pointed out that few-shot LLMs show limited performance on specific NLP tasks \cite{ma-etal-2023-large, zhang2023aligning}. \citet{ma-etal-2023-large} concluded that LLMs are not good at IE tasks such as NER, RE, and event argument extraction. Therefore, we also expect that the LLM-generated data for SFA will have limited accuracy compared to human-annotated data. 

% Many studies have also pointed out that few-shot LLMs often underperform compared to SLMs (\citealp{ma-etal-2023-large}; \citealp{zhang2023aligning}). \citet{ma-etal-2023-large} found that when sufficient training data is available (e.g., hundreds of training samples), SLMs outperform few-shot LLMs in information extraction (IE) tasks. This not only indicates that LLMs may\citet{wang-etal-2021-want-reduce} also observed that a smaller in-house model matches or even exceeds the performance of an LLM, even when the training labels are generated by the LLM. Furthermore, LLMs have been noted to have higher output latency and deployment costs compared to fine-tuned SLMs. Therefore, a smaller in-house SLM is still desirable in situations where inference time, running cost, and privacy concerns are the main obstacles. 

% Some studies have attempted to enhance the limited performance of LLMs on information extraction (IE) tasks by augmenting the few-shot selection process \citep{wan2023gptre}, combining the strengths of LLMs and SLMs \citep{ma-etal-2023-large}, adding an extra layer of explanation before annotation \citep{he2023annollm}, or adjusting prompts and option choices to more closely align with the instruction training data format \citep{zhang2023aligning}. 

\textbf{LLMs as Data Annotators.} There have been several efforts to generate synthetic data from LLMs to train SLMs, primarily to maintain privacy and reduce costs. \citet{wang-etal-2021-want-reduce} utilized few-shot GPT-3 to generate labels for natural language understanding and natural language generation tasks, achieving performance comparable to human labeling while significantly reducing costs. \citet{ding2023gpt3} explored various methodologies to generate labeled data using GPT-3, fine-tuning an SLM that performed comparably to a model trained on human-labeled data in tasks such as sentiment triplet extraction. However, existing LLM-as-annotators approaches have only explored sentence-level labels or relation triplets, and thus do not target tasks like SFA that require sequence-labeling outputs to handle entities scattered across utterances. Moreover, they do not provide a comprehensive analysis on how to allocate the budget between human and LLM-generated data.
% They specifically mentioned that with a limited budget, using LLM-generated data could achieve the same level of performance as human-labeled data, matching the raw performance of the LLM. 
% \citet{josifoski-etal-2023-exploiting} enhanced the quality of LLM-generated data for various IE tasks by exploiting the asymmetry in synthetic data generation, sampling true triplet sets first, and then generating texts containing those triplets. 

% combined to the end of ↑
% However, it is worth mentioning that these studies do not focus on tasks like SFA that require its output to be in sequence labeling format to process details scattered across multiple sentences, and nor do they provide a comprehensive analysis of how to distribute the budget between LLM-generated and human data. To our knowledge, previous methods could only generate sequence-level labels or relation triplets from an LLM; therefore, it is difficult or impossible to convert to sequence labeling style data when there are many types of labels in the sequence.

\section{Preliminaries}
We define semantic frame analysis (SFA) and introduce the EIDC dataset we used in this study. 

\subsection{Semantic Frame Analysis (SFA)}
Semantic frame analysis aims to extract semantic frames, which represent events, in a given context. The core of a semantic frame is a \textbf{trigger}, which is a predicate and the main action of the event. Since each frame has only one trigger, we refer to the frame type by the trigger type from now on without further notice. The event can also include associated details, such as the object, instrument, or temperature, referred to as frame \textbf{arguments}, linked to the event-evoking trigger. Note that different from frame designs such as the FrameNet project \citep{baker-etal-1998-berkeley-framenet}, all frames share common argument types in the EIDC dataset.
% Note that one can choose to adopt existing frame definitions, for example, the frame designs from FrameNet \citep{baker-etal-1998-berkeley-framenet}, or to design frame structures that suit their own need, which is the case in the EIDC dataset. In FrameNet, different frame types can have different argument roles, but in the EIDC setting, all frames share common argument types. 

% There are two methods for conducting SFA: extracting only event-evoking triggers and argument triplets, or identifying triggers and arguments using a sequence-labeling approach that considers entity relations. For technical interview dialogues, a model must process multiple speaker utterances, and the conversational nature means the same events and entities are frequently referenced. Therefore, this work adopts the second method, solving SFA through sequence labeling to capture entity token positions. A demonstration of sequence labeling is provided in Figure \ref{fig:dialogue_example}.

SFA consists of two parts: \textbf{Trigger Detection} and \textbf{Argument Detection}. In this work, we formulate SFA as a sequence labeling task to capture entities scattered across multiple utterances. Therefore, detection means identifying both the span and the type of an entity. In addition to entity span and type, an argument must link to a detected trigger. A visual example of SFA annotation is presented in Figure~\ref{fig:dialogue_example}. We designed a novel prompting and output format for an LLM to handle SFA efficiently (Section~\ref{synth_label}), and utilized an architecture that can handle both sequence labeling and relation extraction at the same time (Section~\ref{jamie}). 

% Since we are conducting SFA in a dialogue scenario, there are expected to be multiple utterances and many entities in a single context. Therefore, both the LLM and the SLM should be able to handle entity spans within the context. For the output format of the LLM, refer to Section \ref{synth_label}. For the SLM, we used a model that can handle both entity spans and relation extraction (Section \ref{jamie}).

\subsection{Technical Interview Dialogue Dataset with SFA Annotation}
In this paper, we utilize the \textit{cooking} section of the EIDC dataset \cite{okahisa_2022_constructing, chika_2024_domain}. Note that from now on, by the EIDC dataset, we refer to the cooking section without further notice. The dataset is comprised of technical interview dialogues with SFA annotations. % The method for manually collecting the dialogue and conducting the annotation is explained below.

% CIDC (cooking dataset: https://aclanthology.org/2022.lrec-1.335.pdf)の論文を参考にした
\textbf{Technical Interview Dialogues.} \label{manual_dialogue_collection}
The EIDC dataset contains interview dialogues where an expert discusses cooking processes with an interviewer. The expert introduces and explains a recipe spontaneously or in response to the interviewer’s questions. The interviewer is asked to actively elicit knowledge about the cooking process through interactions, such as asking questions.

\textbf{Annotation for Semantic Frame Analysis.}
Each dialogue in the EIDC dataset comes with manual annotations of SFA. Since these dialogues pertain to the cooking domain, the semantic frames are designed to capture cooking-related events. For example in Figure~\ref{fig:dialogue_example}, when a speaker mentions the action of lining something up, the predicate of this event will be annotated with a "PLACE" type of trigger. If the event also involves an object being lined up, that object will be annotated as an "Object" type of argument, and linked to the trigger. The complete list of trigger and argument types can be found in Appendix~\ref{sec:label_dist}.

\section{Data Synthesis With an LLM}
This section presents our methodology for constructing training data for conversational semantic frame analysis using an LLM, as illustrated in the overview in Figure~\ref{fig:overview}. We use an LLM to label either human dialogues or pseudo-dialogues also generated by an LLM, resulting in 2 pseudo-data variants: Human-Pseudo and Pseudo-Pseudo.

\begin{figure}[h]
  % \vspace{-3mm}
  \centering
  \includegraphics[width=0.9\columnwidth]{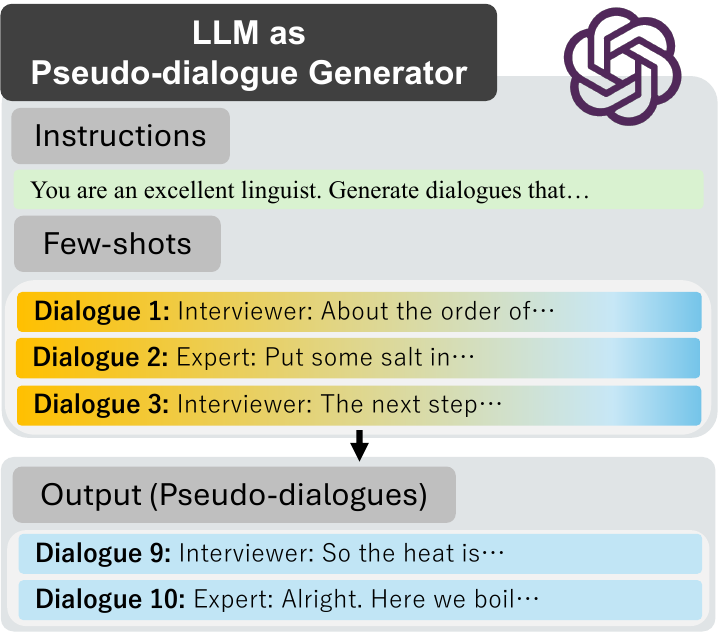}
  \caption {GPT-4 is used as a pseudo-dialogue generator by taking preserved and previously generated dialogue sessions as few-shots. The orange-blue rainbow color indicates that the few-shots contain both human and pseudo-dialogues. Refer to the actual prompt design in Appendix~\ref{sec:pd_prompt_appendix}. }
  \label{fig:pd_generator}
  \vspace{-5mm}
\end{figure}

\subsection{Pseudo-dialogue Generation}\label{pd_generation}
To generate pseudo-dialogues, the LLM is prompted with few-shot dialogues and asked to generate new ones that are close to the few-shots in format but contain different contents (Figure~\ref{fig:pd_generator}). For the few-shot examples, we not only sample from a preserved pool of human dialogues but also adopt the self-instruct strategy \cite{wang_2023_selfinstruct} to sample from the previously generated pseudo-dialogues to increase diversity. The pre-filtering and post-filtering methods, along with the detailed settings for the self-instruction of pseudo-dialogues, are explained in Section~\ref{pd_generation_details}.

\begin{figure}[h]
  \centering
  % \vspace{-5mm}
  \includegraphics[width=0.96\columnwidth]{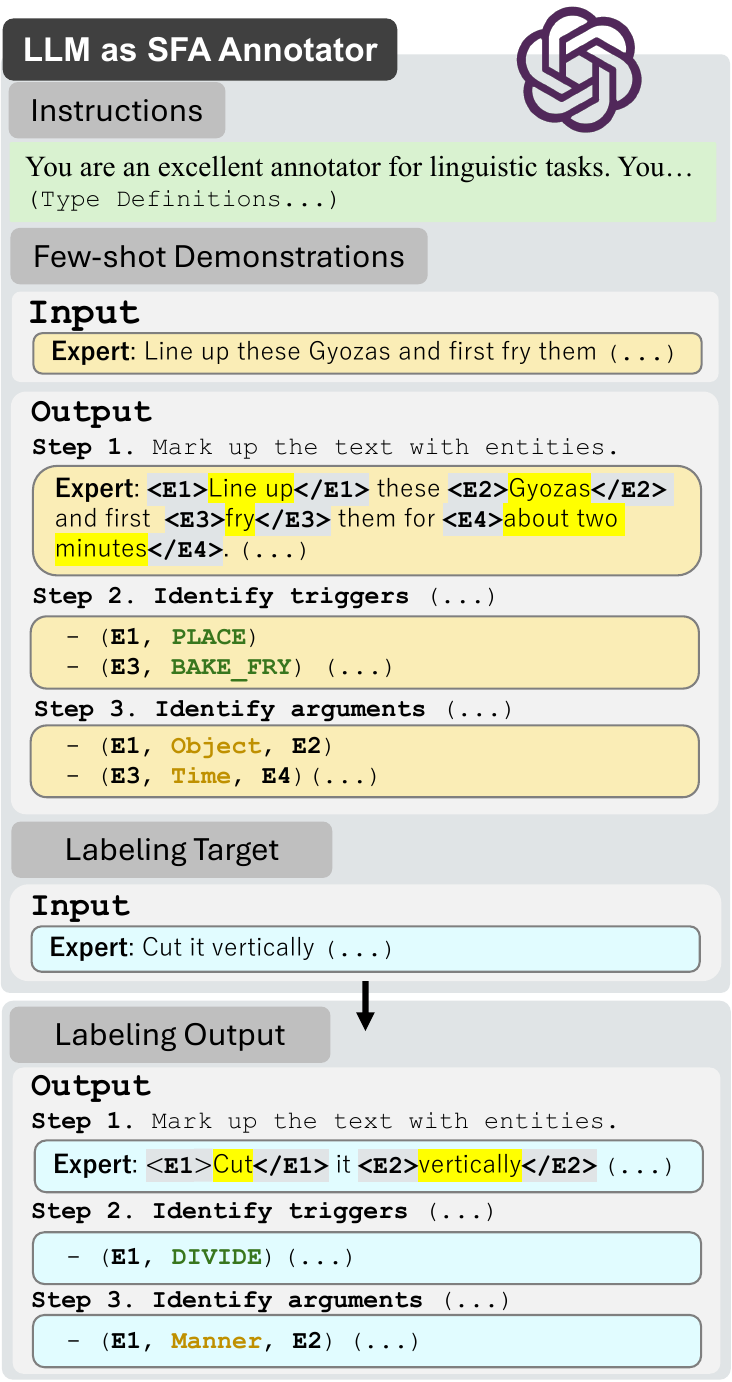}
  \caption {We designed a novel multi-step labeling scheme for LLMs to handle SFA in text generation. Refer to the full prompt design in Appendix~\ref{sec:sfa_prompt_appendix}. }
  \vspace{-5mm}
  \label{fig:sfa}
\end{figure}

\subsection{Pseudo-labels by LLM}\label{synth_label}
We design a novel multi-step labeling approach to convert SFA into a text generation task that can be efficiently managed by an LLM. An example of this pseudo-labeling process is illustrated in Figure~\ref{fig:sfa}. The system prompt includes definitions of trigger and argument types as specified in the annotation guidelines, along with few-shot examples to demonstrate the SFA process in a text generation format. In each example, such as the one in Figure~\ref{fig:sfa}, entities like "line up" and "Gyozas" are tagged with entity tags such as "<E1>" in the first step.
In step 2, the LLM identifies all triggers within these entities. Finally, in step 3, the arguments for each trigger are determined using relation triplets. The output can then be seamlessly converted into sequence labeling data for our SLM.

\subsection{Data Variants}
We construct three data variants with the dialogues and annotations sources from either human or LLM: Human-Human, Human-Pseudo, and Pseudo-Pseudo. In this context, "Human" refers to data collected from humans, while "Pseudo" denotes data generated by an LLM. We did not consider a Pseudo-Human variant because human annotation is too precious to be assigned to lower-quality LLM-generated dialogues. 
% human annotation is costly. Our initial priority was to ensure that pseudo-dialogues do not significantly limit performance on the downstream task.

\textbf{Human-Pseudo.} In this data variant, SFA labels are assigned by an LLM to human dialogues sampled from the EIDC dataset. This setting reflects the scenario where one has already collected the text part of their data and has started to apply labels for their task. 

\textbf{Pseudo-Pseudo} This is a fully synthesized data variant with LLM-generated dialogues and labels. This variant is the cheapest and the least time-consuming, as you only need some few-shot examples to start crafting data. 

\textbf{Human-Human.} We sampled human dialogues and labels directly from the EIDC dataset and formed Human-Human data. The Human-Human data is the most expensive and is also expected to have the highest label accuracy, closely aligning with the desired standards defined in the annotation guidelines.

\begin{table*}[h]
  \centering
  \begin{tabular}{c|r|rrr}
    \toprule
     & \textbf{Data Size}& \multicolumn{3}{c}{\textbf{Cost}} \\
    \cline{3-5}
    \textbf{Data Type} & \textbf{(Sessions)} & \textbf{Text (\$)}& \textbf{Label (\$)}& \textbf{Total (\$)} \\
    \hline
    Human-Human     & 1,472 & 6.4k & 6.4k & 12.8k          \\
    \hline
    Human-Pseudo     & 2,858 & 12.4k & 0.37k & 12.8k            \\
    Pseudo-Pseudo     & 4,293 & 0.28k & 0.56k & 0.84k           \\
    % \hline
    % Valid     & 269 &  &  &            \\
    % Test      & 379 &  &  &             \\
    \bottomrule
  \end{tabular}
  \caption{The size and cost statistics of the three data variants.}
  % \caption{The size and cost of the three data variants. Note that the full cooking section of the EIDC dataset contains 4,600 dialogue sessions, from which we sampled human dialogues for the Human-Pseudo data. The cost for Human-Human data is estimated using the total cost of that: \$40,000. }
  \label{tab:size_and_cost_stats}
  \vspace{-4mm}
\end{table*}

\section{Experiments}
To investigate how LLM-generated data can contribute to optimal cost-efficiency, we first defined the budget ranges and assembled both human and LLM-generated data according to these budget settings. From the EIDC dataset, we sampled up to \$12,800 to create the Human-Human data. We then synthesized two types of LLM-generated data: Human-Pseudo for \$12,800 and Pseudo-Pseudo for \$840. Finally, we investigated the optimal ratio for combining Human-Human data with LLM-generated data under each budget to achieve the best SFA performance. To do this, we trained an SLM using different data combinations and evaluated its performance based on trigger detection and argument detection. The following subsections provide detailed descriptions of the experimental process, results, and analyses.
Note that to fit within the context length limits of both the LLM and our SLM, we divide dialogues into smaller sessions using a heuristic method. Hereafter, a 'dialogue' will refer to a 'dialogue session' unless otherwise specified. Each session typically consists of about 10 utterances. 
% We provide details on our model settings. Note that to fit into the context length limit of both the LLM and our SLM, we divide the dialogues into smaller sessions using a heuristic method. In the following context, a "dialogue" refers to a "dialogue session" without further notice.

% \subsection{LLM Settings for Pseudo-dialogue Generation} \label{pd_generation_details}
\subsection{Details of Data Synthesis Procedures} \label{pd_generation_details}
% We introduce detailed settings of the two main components used to synthesize Human-Pseudo and Pseudo-Pseudo data: the pseudo-dialogue generator and the pseudo-labeler, both powered by GPT-4.

\textbf{Pseudo Dialogue Generator.} As introduced in Section~\ref{pd_generation}, we adopted the self-instruct strategy \cite{wang_2023_selfinstruct} to bootstrap pseudo-dialogue generation. Mostly following the settings in their work, we provide the model with 8 dialogues as few-shots: 6 human dialogues and 2 pseudo-dialogues for topic diversity. Since we did not have pseudo-dialogues when we started, we first created a few pseudo-dialogues with few-shot examples containing only human dialogues. Afterward, we moved on to mixing few-shot examples. Before adding pseudo-dialogues back into the dialogue pool, we filtered them by ROUGE-L score (<0.7) against existing dialogues to ensure that the newly generated ones were not extremely similar to the existing ones. None of the pseudo-dialogues exceeded this limit. We then filtered the most similar ones using ROUGE-L to reduce them to the desired size shown in Table~\ref{tab:size_and_cost_stats}, which ended with a max ROUGE-L score of 0.52. We used GPT-4-0613 (accessed 01/2024) and set the generation temperature to 0.7.

% \subsection{LLM Settings for SFA Labeling}
\textbf{Pseudo SFA Labeler.} \label{llm-instability}
We adopted GPT-4-0613 (accessed 01/2024) to generate pseudo-labels for SFA. For few-shots, we sampled 3 complete human dialogues, then filtered them to remove sessions with too few entities, resulting in 37 dialogue sessions. For each labeling target, we used 3 few-shots: the top 2 most similar dialogue sessions, determined by the ROUGE-L score to ensure similarity to the target, and 1 specially preserved dialogue session containing as many as 30 entities. This special few-shot was included in all cases because we empirically observed that GPT-4 tends to overlook entities if the few-shots lack sufficient entities.\footnote{We also observed that GPT-4 sometimes violated the instructions by altering the context or refusing to label. For less powerful LLMs such as GPT-3.5 Turbo and GPT-4 Turbo (GPT-4-1106-preview), this problem was even more severe and made them unusable.} 

\begin{figure*}[t]
  \vspace{-4mm}
  \centering
  \includegraphics[width=0.49\linewidth]{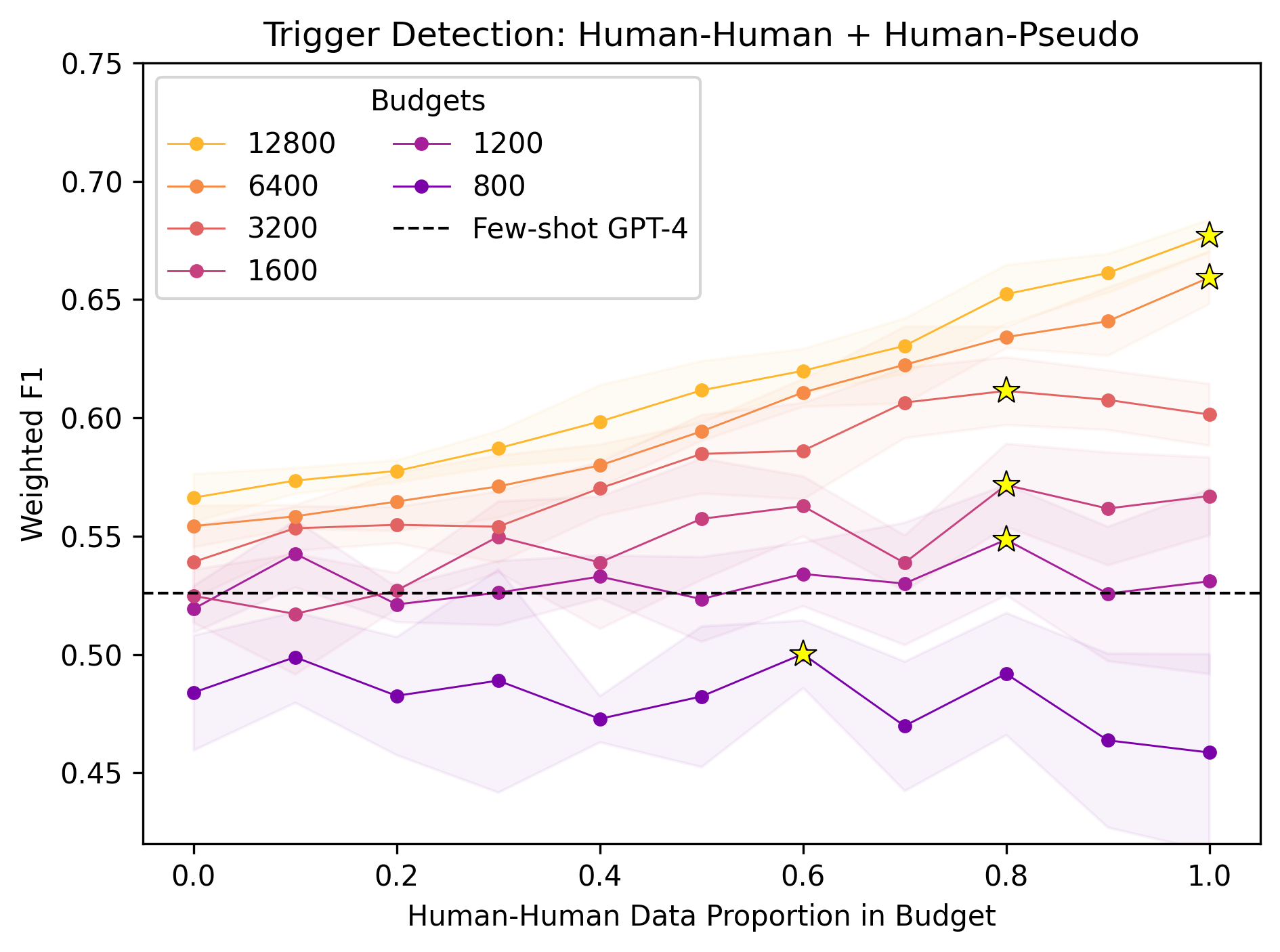} 
  % \hfill
  \includegraphics[width=0.49\linewidth]{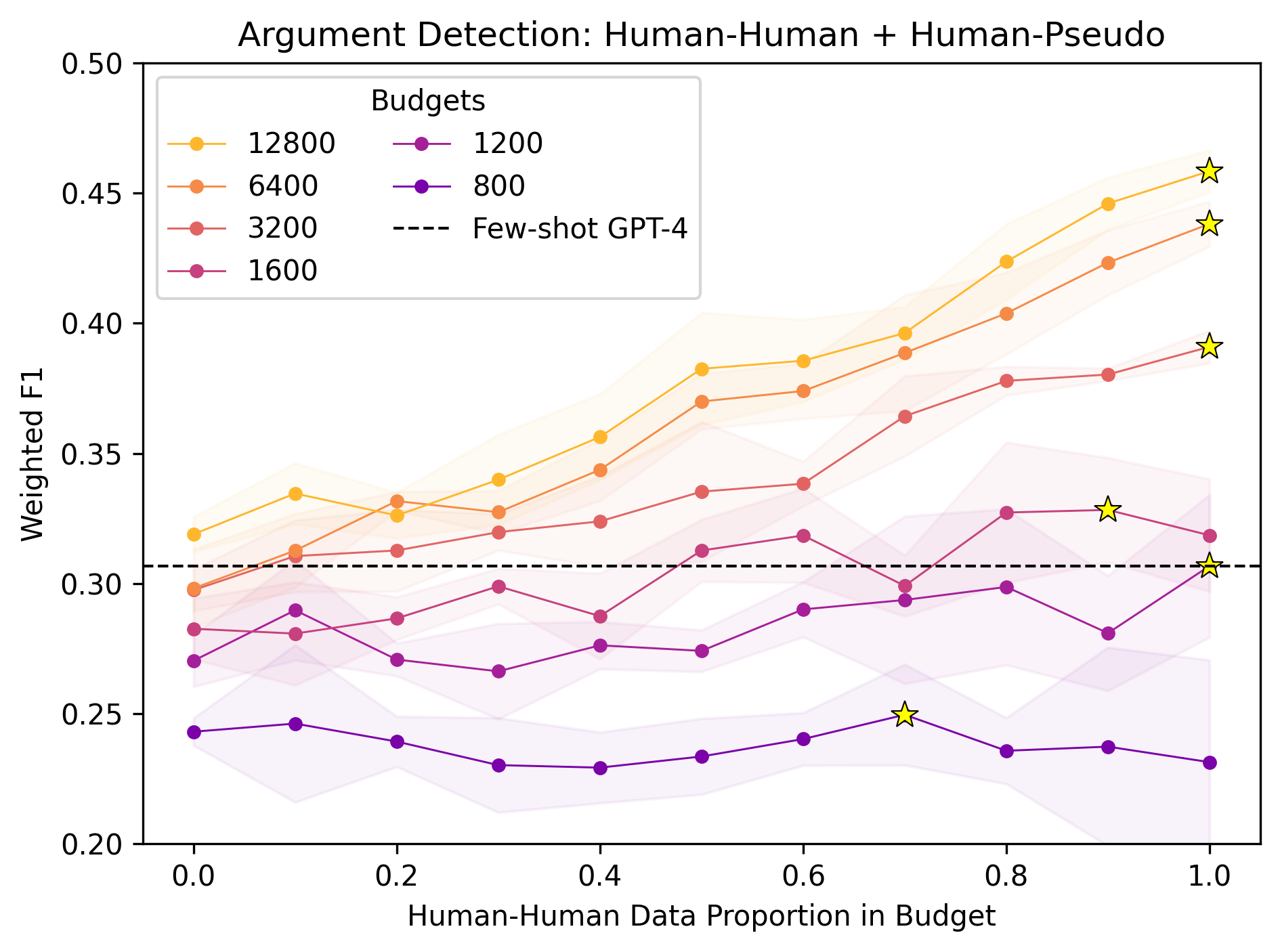}
  \vspace{-3mm}
  \caption {The budget-wise cost-efficiency plot for combining Human-Human and Human-Pseudo data. The black dotted line represents the performance of few-shot GPT-4. Each budget curve features a star marking its optimal point. The shaded region around each curve indicates the standard deviation across five different seeds.}
  \vspace{-3mm}
  \label{fig:rp}
\end{figure*}

\begin{figure*}[t]
  \centering
  \includegraphics[width=0.49\linewidth]{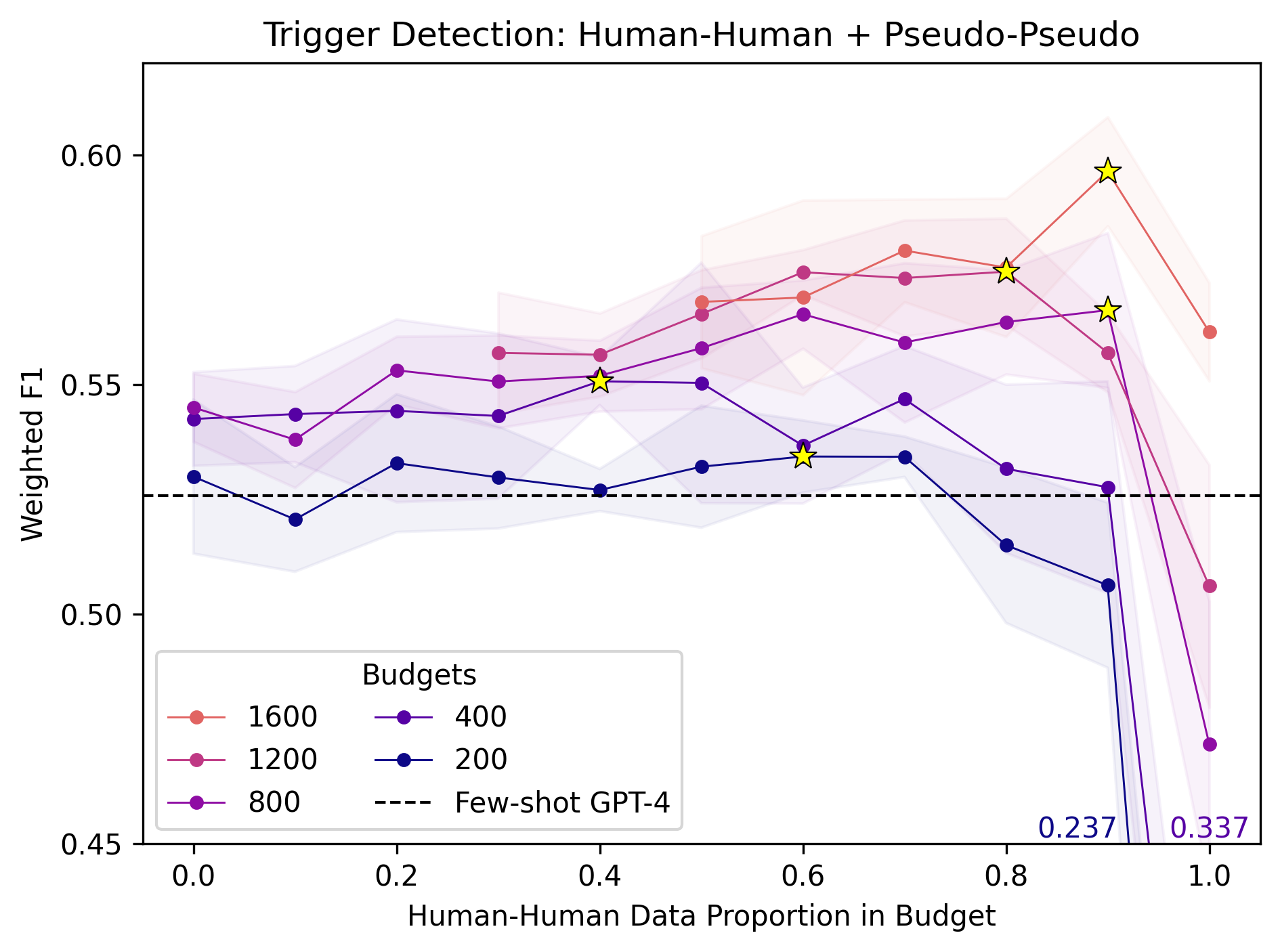} 
  % \hfill
  \includegraphics[width=0.49\linewidth]{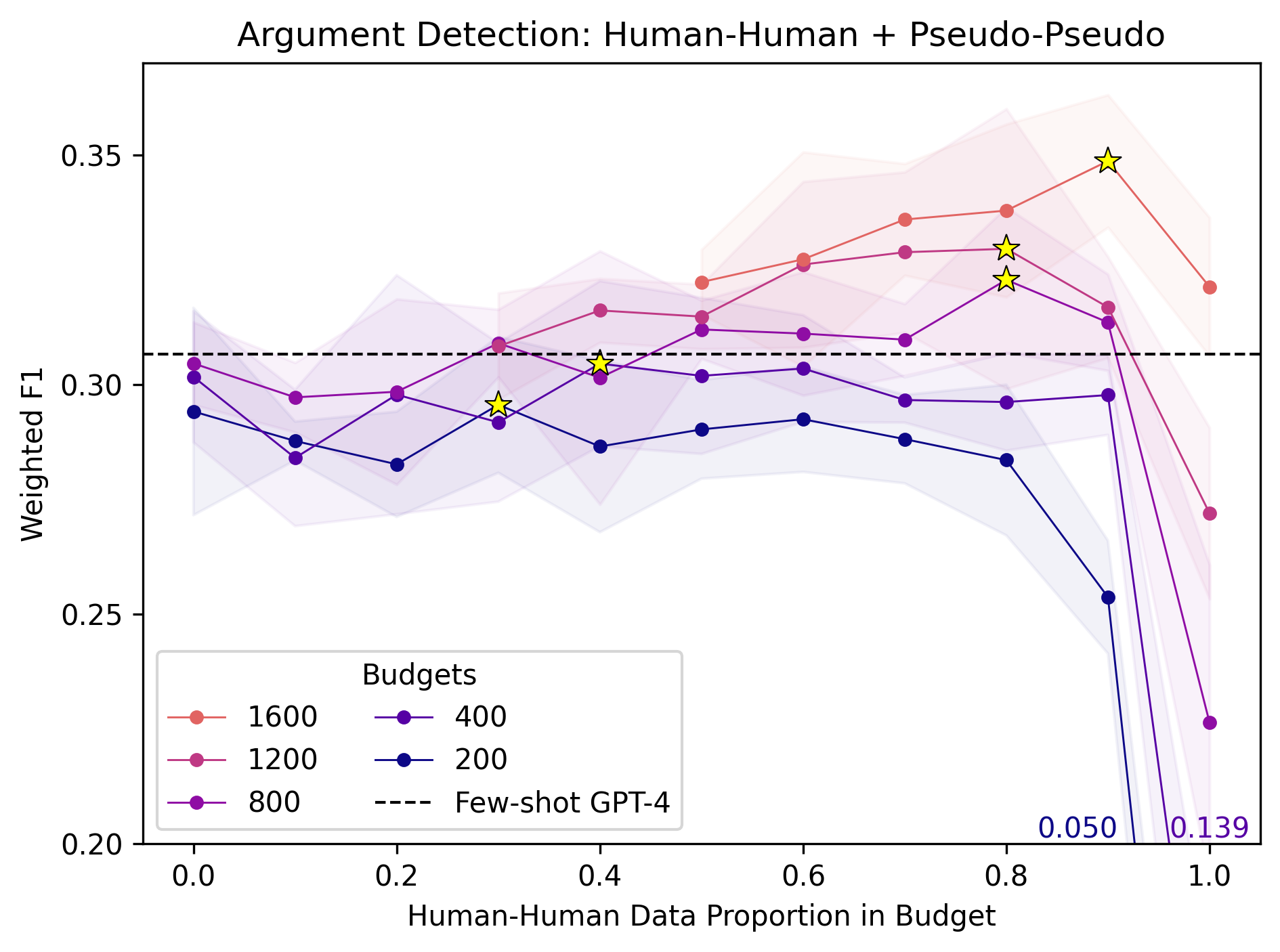}
  \vspace{-3mm}
  \caption {The budget-wise cost-efficiency plot for combining Human-Human and Pseudo-Pseudo data. Due to the collection limit of \$840 worth of Pseudo-Pseudo data, the plot only shows the right portion of the curve for budgets of \$1,200 and \$1,600, where the data is combined with Human-Human data. The values of some out-of-range data points are displayed on the plot with colors corresponding to the budget curve. }
  \label{fig:pp}
  \vspace{-3mm}
\end{figure*}

\subsection{Budget Settings and Data Statistics}
We provide detailed information on the budget settings, costs, and basic statistics for the three types of data variants: Human-Human, Human-Pseudo, and Pseudo-Pseudo.

\textbf{Total Data Sizes and Costs.\footnote{We excluded the collection cost of few-shot examples sampled from the training split of the EIDC dataset, as well as the instructions derived from the annotation guidelines. }} 
% Setting an adequate budget range is crucial for analyzing the cost-efficiency of the data. The budget range should reflect both the high accuracy of human-annotated data and the low cost of LLM-generated data. 
As shown in Table~\ref{tab:size_and_cost_stats}, we collected up to \$12,800 for both Human-Human and Human-Pseudo data, which roughly aligns with the three-year total of scholarship funds for a PhD student at a Japanese university. For Human-Human data, we extracted \$12,800 worth of human dialogue and label pairs from the EIDC dataset, out of a maximum of 4,600 sessions and a total cost of \$40,000 of the original dataset. In the EIDC dataset, the costs for human dialogues and human labels are roughly the same. For Human-Pseudo data, we repeatedly applied pseudo-labels to the existing human dialogues in the EIDC dataset until the cost reached \$12,800, which was calculated based on the cumulative costs of the human dialogues and OpenAI API usage. Notably, the pseudo-labels accounted for only 3\% of the total cost of Human-Pseudo data. As a result, we were able to annotate more dialogues than the Human-Human data. For Pseudo-Pseudo data, due to the low cost of both pseudo-dialogue and pseudo-labels, we collected 1.5x times the data size compared to Human-Pseudo data while only costing \$840. % The cost for pseudo-dialogues takes up $\frac{1}{3}$ of \$840, while pseudo-labels account for the remaining $\frac{2}{3}$. 
The costs for pseudo-dialogues and pseudo-labels were calculated from the token usage of the OpenAI API service.
We ceased further collection of Pseudo-Pseudo data upon discovering that performance had reached saturation and would not improve with additional data.

% We mix Human-Human data with Human-Pseudo or Pseudo-Pseudo data and adjust the ratio under each fixed budget to identify the combination that leads to optimal cost-efficiency. For example, in Figure \ref{fig:rp}, with a budget of \$12,800, we start with all Human-Human data in the mix and plot a performance data point on the top right. Then, we gradually increase the Human-Pseudo data in the mix while decreasing the amount of Human-Human data, plotting more data points towards the left. Eventually, we reach a mix with full Human-Pseudo data and no Human-Human data, and plot the final data point on the left. During this process, we only adjust the amounts of the two types of data while keeping the total budget fixed at \$12,800.

\textbf{Budget Setting for Experiments. }We set a series of budgets of \$800, \$1,200, \$1,600, \$3,200, \$6,400, \$12,800 for Human-Human and Human-Pseudo mixture, and \$200, \$400, \$800, \$1,200, \$1,600 for Human-Human and Pseudo-Pseudo mixture. For each budget, we adjust the budget proportion of Human-Human data from 0 to 1 with an interval of 0.1.  
% Although \$1,200 is not half of \$3,200, we included it to gather more detailed information. 

 % For example, in Figure \ref{fig:rp}, the x-axis is the proportion of Human-Human data within the mixture, so the points in the middle mean a 1 to 1 (0.5:0.5) ratio of Human-Human and Human-Pseudo data.

\textbf{Length and Label Distributions in Dialogues.}
We conducted a quantitative analysis comparing human dialogues and pseudo-dialogues. We found that the average length of pseudo-dialogues generated by GPT-4 was similar to that of human dialogues (127 tokens vs. 136 tokens) and exhibited fewer extreme outliers in terms of length. By comparing the label density of Human-Pseudo and Pseudo-Pseudo data, we observed that pseudo-dialogues tended to contain more entities than human dialogues, leading to a higher count for certain label types. For more details on the length and label distributions of pseudo-dialogues, refer to Appendix~\ref{sec:length_dist} and Appendix~\ref{sec:label_dist}.

\subsection{SLM and Evaluation Metrics for SFA}

% \textbf{SLM for SFA.} 
\label{jamie}
We adopt JaMIE \citep{cheng-etal-2022-jamie} as our SLM for SFA. JaMIE is an architecture featuring one encoder and multiple decoders for sequence labeling and can handle relation extraction by design. 
%The first decoder identifies whether a token is a trigger, a non-trigger entity, or a non-entity. The second decoder further identifies the type of triggers. The final decoder determines the relationships between entities, such as identifying the arguments of a trigger.
We employ the Japanese DeBERTa-V2-base as the pre-trained encoder for JaMIE and train the relation decoders from scratch.\footnote{\href{https://huggingface.co/ku-nlp/deberta-v2-base-japanese}{https://huggingface.co/ku-nlp/deberta-v2-base-japanese}} Refer to the training hyperparameters in Appendix~\ref{sec:hyperparams}.

We evaluate the performance of SFA using a classification metric with a special focus on entity spans. Both the type and the span of the entity should be correct to be counted as correct. Partial scores are awarded if the span overlaps with the true label.\footnote{We modified the evaluation code from seqeval (\href{https://github.com/chakki-works/seqeval}{https://github.com/chakki-works/seqeval}).} For arguments, since they rely on a trigger, only those whose target trigger is predicted correctly are counted as correct.\footnote{In addition to semantic frames, the data also includes Event Coreference Relations (ECR). We did not evaluate ECR directly, however, we evaluated argument detection by allowing the target trigger to be any of the events on the same ECR event sequence in the true labels.} We calculate each class's F1 score and derive a weighted F1 score for triggers and arguments, respectively, where the weights are calculated based on the number of instances in each class.

% Note that in addition to semantic frames, our data also includes Event Coreference Resolution (ECR) relations, which aim to determine if one trigger represents the same event as another, for both the LLM and the in-house JaMIE model. However, we did not evaluate ECR relations in this study.

% The weighted F1 score is defined as follows:
% \begin{equation}
% \textrm{Weighted F1} = \sum_{i=1}^{n} w_i \cdot \textrm{F1}_i
% \end{equation}
% where
% \begin{equation}
% w_i = \frac{s_i}{\sum_{i=1}^{n} s_i}
% \end{equation}

% $s_i$ is the number of instances of class $i$.

\subsection{Main Results}
We report on the cost-efficiency of incorporating Human-Pseudo and Pseudo-Pseudo data.
% There are different colors of curves on each diagram, each curve stands for a specific amount of budget. The x-axis of the cost-efficiency diagram is the proportion of Human-Human data. Therefore, data points on the very left use 100\% LLM-generated data. Moving on to the right, the proportion of Human-Human data gradually increases, and 100\% of Human-Human data is used on the very right. The y-axis shows the weighted-f1 score. 

\textbf{Incorporating Human-Pseudo Data.} \label{HP_data}
% \textbf{Using one type of data only. } 
% In each plot, the data points on the far left of the plot are denser than those on the right. We believe this is because Human-Pseudo and Pseudo-Pseudo data are cheaper and reach performance saturation more quickly, while the gold data is more expensive but has greater potential for performance improvement.
In Figure~\ref{fig:rp}, we observe that when the budget is lower than \$6,400 for trigger detection and \$3,200 for argument detection, optimal cost-efficiency is achieved by combining Human-Human and Human-Pseudo data. The lower the budget is, the more Human-Pseudo data should be included for best performance. In this case, the trade-off between human data and LLM-generated data has a positive impact on the performance. 

On the other hand, we see that when the budget is higher than above, the optimal cost-efficiency is brought by using 100\% Human-Human data. This shows that LLM-generated data cannot be used in all situations because it may harm the performance due to its lower accuracy.

% In other words, this is a trade-off between higher accuracy but lower quantity of the Human-Human data, and the lower accuracy but greater quantity of the Human-Pseudo data.
% We believe that the optimal point appears in the middle of these budget curves because:
% \begin{enumerate}
%     \item Starting from the left and moving to the right: incorporating Human-Human data helps improve performance since Human-Human data is more accurate.
%     \item Starting from the right and moving to the left: trading expensive Human-Human data for a greater quantity of LLM-generated data is effective when performance has not yet reached saturation.
% \end{enumerate}

% When the budget is extremely low, we observed that the cost-efficiency curve shows a downward trend toward the right. This indicates that Human-Pseudo data is very effective due to its low cost in these situations. 

\textbf{Incorporating Pseudo-Pseudo Data.}
In Figure~\ref{fig:pp}, we see that for all the budgets we set, the optimal performance was achieved by combining Human-Human and Pseudo-Pseudo data. We specifically observed that since Pseudo-Pseudo data is so much cheaper, allocating 10\% of the budget to Pseudo-Pseudo data in budget \$1,600 brought a significant performance boost for both trigger and argument detection. Although we did not further raise the budget for Pseudo-Pseudo data, we can estimate that the optimal will be achieved by using 100\% Human-Human data if we raise the budget to \$6,400 and above. Therefore, we draw a similar conclusion as in Human-Pseudo data: when your budget is not high enough to reach saturation (optimal performance by 100\% Human-Human data), one should incorporate Pseudo-Pseudo data to achieve optimal cost-efficiency. 
% Furthermore, Pseudo-Pseudo data is even more effective when the budget is extremely low, and one can consider using 100\% of Pseudo-Pseudo data in these situations.

\subsection{Findings}
We further investigated whether Pseudo-Pseudo data is inferior to Human-Pseudo data because of the pseudo-dialogues it contains. Additionally, we evaluated the effectiveness of LLM-generated data from a data augmentation perspective.

\textbf{Human-Pseudo vs. Pseudo-Pseudo.} We observed no significant disadvantage caused by replacing human dialogues with pseudo-dialogues in the training data. With the same budget of \$1,600, one could achieve a slightly higher performance in trigger detection using Pseudo-Pseudo data compared to Human-Pseudo data (0.596 vs. 0.571). Additionally, by comparing the data points using all LLM-generated data in both plots, we noticed that Pseudo-Pseudo data achieves the same level of performance while costing about $\frac{1}{10}$ of Human-Pseudo data (\$200 vs. \$1,600 in trigger detection). 
 
% Therefore, we conclude that it is advisable to generate fully synthesized data like the Pseudo-Pseudo data for SFA when labeling accuracy is the main bottleneck of LLM-labelers. 

\begin{figure}[h!]
  \vspace{-2mm}
  \centering
  \includegraphics[width=0.97\columnwidth]{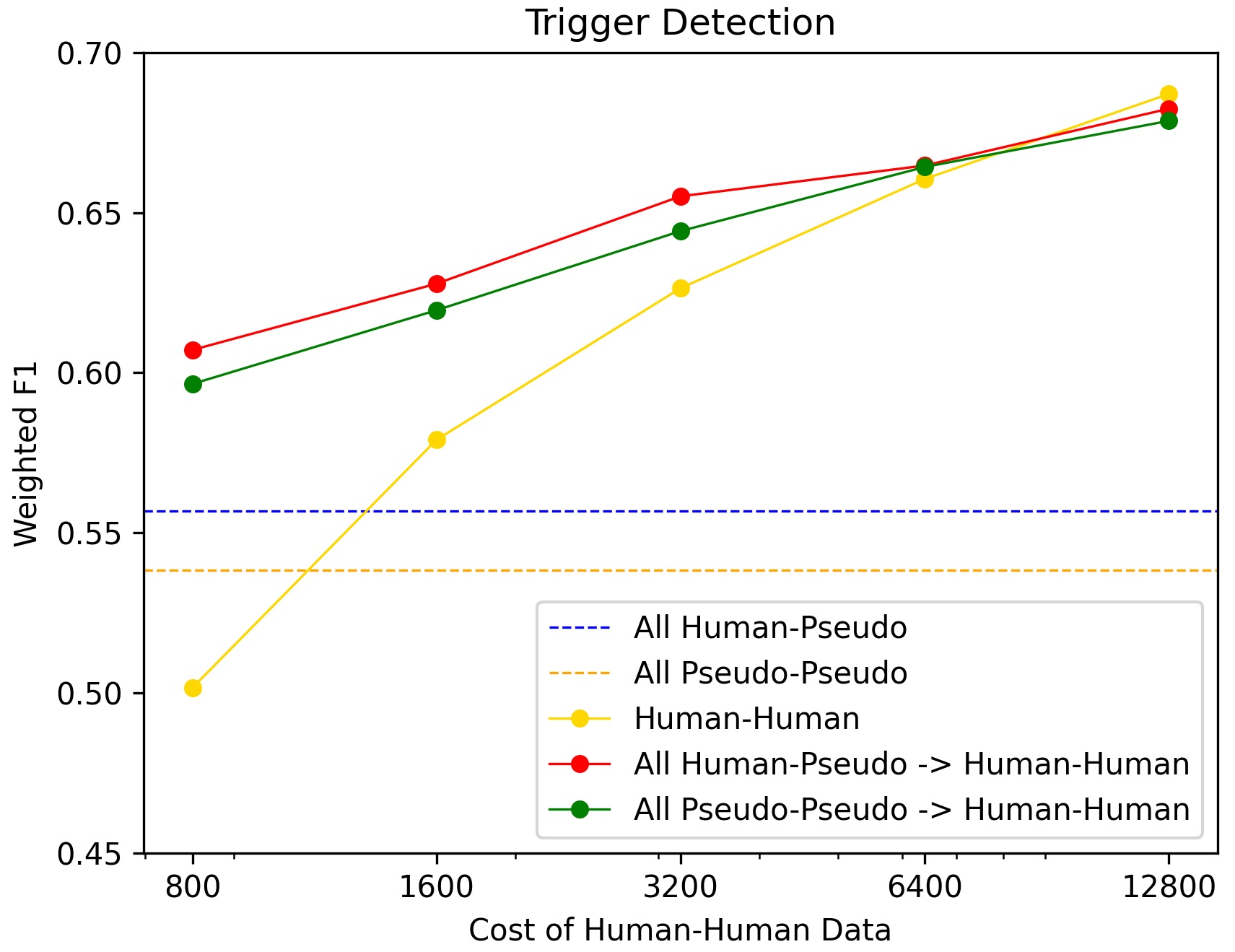} \\
  \vspace{1mm}
  \includegraphics[width=0.96\columnwidth]{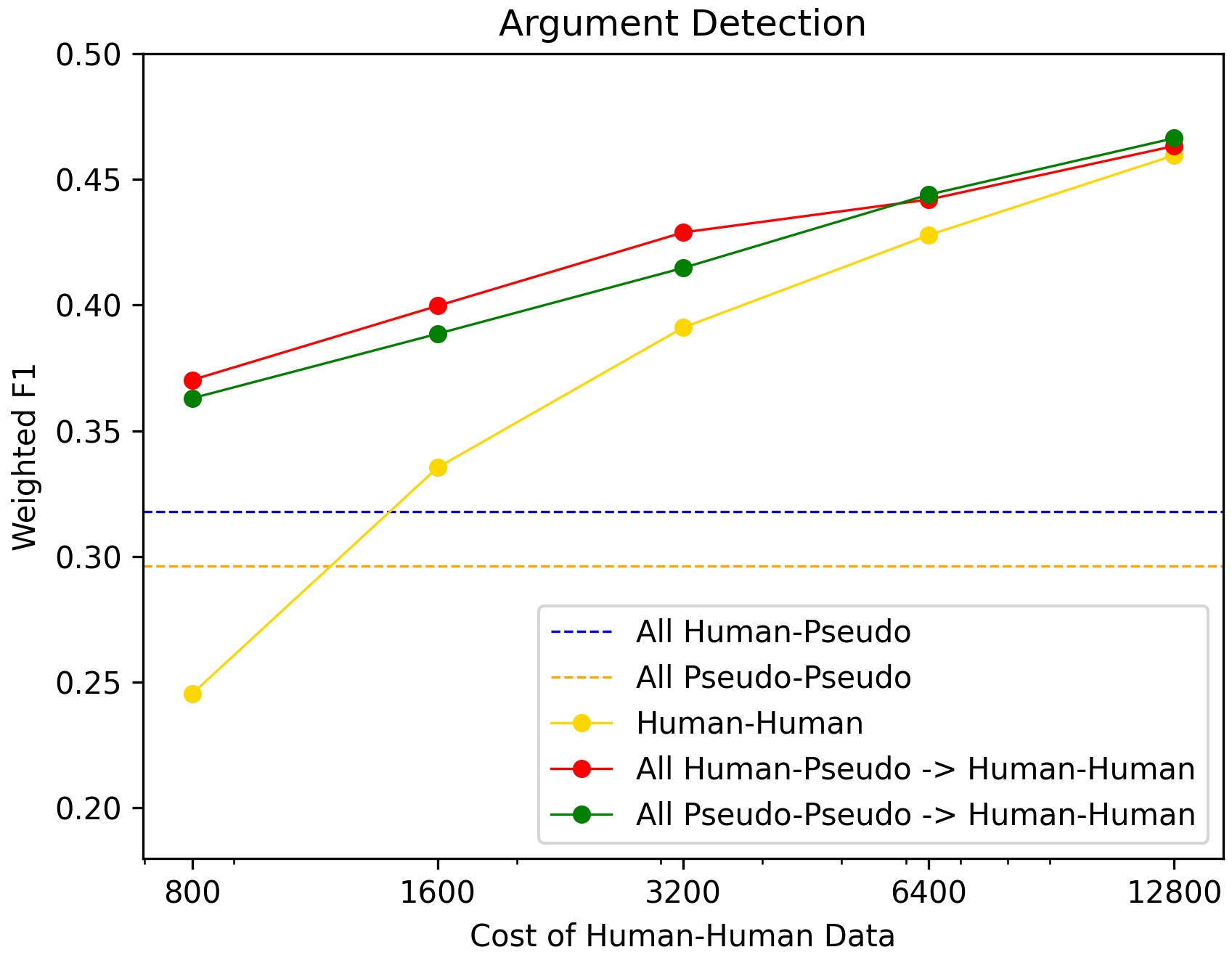}
  \caption {The effectiveness of LLM-generated data from a data augmentation perspective. We trained the SLM on all Human-Pseudo or Pseudo-Pseudo data (blue and orange dotted lines), then continued training on different sizes of Human-Human data (red and green lines). }
  \label{fig:da}
  % \vspace{-5mm}
\end{figure}

\textbf{From a Data Augmentation Perspective.}
We review the effectiveness of LLM-generated data from a data augmentation perspective (Figure~\ref{fig:da}). In this setting, we trained the SLM first using all LLM-generated data, i.e., either all Human-Pseudo or Pseudo-Pseudo data, then continued training it on different costs of Human-Human data, ranging from \$800 to \$12,800. The result shows that when the amount of Human-Human data is limited (lower than \$3,200), both Human-Pseudo and Pseudo-Pseudo data help boost performance. The effectiveness of LLM-generated data is more significant when the budget for Human-Human data is low. Notably, while the cost of Pseudo-Pseudo data is significantly cheaper than Human-Pseudo data in this setting (\$840 vs. \$12,800), the former is arguably competitive against the latter as the max performance gap (green line vs. red line) is less than 0.02 F1 score.

\section{Conclusion}
In this work, we explored the feasibility of using LLM-generated training data for Japanese conversational semantic frame analysis (SFA) and examined its cost-efficiency when combined with human data under various budgets. % We investigated the trade-off between the higher accuracy and cost of human data versus the lower accuracy and cost of LLM-generated data. 
Our findings show that combining both data types is ideal for optimal performance across a wide range of budgets, with more LLM-generated data favored as the budget decreases. Additionally, we compared two variants of LLM-generated data: Human-Pseudo and Pseudo-Pseudo. 
% Both variants have LLM-generated labels, but the former uses human dialogues, while the latter uses LLM-generated dialogues. 
The results indicate that it is viable to use fully synthesized data, i.e. Pseudo-Pseudo, as it significantly lowers the cost to achieve the same level of performance as Human-Pseudo. 
% without compromising downstream task performance when the main bottleneck is the labeling accuracy of the LLM.

In this study, we provided insights specifically on conversational SFA. We believe our conclusions can be extended to similar information extraction tasks such as relation extraction and frame semantic parsing, which future work could explore.

% \begin{itemize}
%     \item We investigated the cost-efficiency of mixing LLM-generated data and human-labeled data to train a smaller supervised model for downstream usage. 

%     \item Experimental results show that LLM-generated data with limited label accuracy is still effective in low-budget settings.

%     \item We provide a method to check when to rely on LLM-generated data and when to switch to full human data.

% \end{itemize}

\newpage

\section{Limitations}
While we believe the conclusions of our work are comprehensive under our settings, there are several limitations. Firstly, we conducted experiments only with GPT-4, as it was the most powerful LLM available at the time and we observed that less powerful LLMs were unable to handle this task, as mentioned in Section~\ref{llm-instability}. Secondly, we did not conduct a qualitative analysis comparing LLM-generated data to human data. 
% For pseudo-labels, we believe that some classes remain difficult for an LLM to distinguish. For pseudo-dialogues, we also believe that there are fundamental differences between pseudo-dialogues and human dialogues. 
Certain aspects of the LLM-generated data, such as the increased entity frequency we observed in the pseudo-dialogues (Appendix~\ref{sec:label_dist}), could indeed affect its effectiveness. 
%Some quality aspects of the LLM-generated data could indeed significantly impact its effectiveness. 
Lastly, estimating the effective budget range for LLM-generated data is not straightforward when adapting to new tasks. The effective range can vary significantly depending on the specific data and tasks involved. We believe future work should explore different LLMs, compare LLM and human data more deeply, and better estimate effective budget range to fully understand the potential and limitations of LLM-generated data.

% \subsection{}

% \section{Acknowledgments}

% \subsection{}

\bibliography{custom}

\begin{thebibliography}{19}
\providecommand{\natexlab}[1]{#1}

\bibitem[{Baker et~al.(1998)Baker, Fillmore, and Lowe}]{baker-etal-1998-berkeley-framenet}
Collin~F. Baker, Charles~J. Fillmore, and John~B. Lowe. 1998.
\newblock \href {https://doi.org/10.3115/980845.980860} {The {B}erkeley {F}rame{N}et project}.
\newblock In \emph{36th Annual Meeting of the Association for Computational Linguistics and 17th International Conference on Computational Linguistics, Volume 1}, pages 86--90, Montreal, Quebec, Canada. Association for Computational Linguistics.

\bibitem[{Brown et~al.(2020)Brown, Mann, Ryder, Subbiah, Kaplan, Dhariwal, Neelakantan, Shyam, Sastry, Askell et~al.}]{brown2020language}
Tom Brown, Benjamin Mann, Nick Ryder, Melanie Subbiah, Jared~D Kaplan, Prafulla Dhariwal, Arvind Neelakantan, Pranav Shyam, Girish Sastry, Amanda Askell, et~al. 2020.
\newblock Language models are few-shot learners.
\newblock \emph{Advances in neural information processing systems}, 33:1877--1901.

\bibitem[{Cheng et~al.(2022)Cheng, Yada, Tanaka, Aramaki, and Kurohashi}]{cheng-etal-2022-jamie}
Fei Cheng, Shuntaro Yada, Ribeka Tanaka, Eiji Aramaki, and Sadao Kurohashi. 2022.
\newblock Jamie: A pipeline japanese medical information extraction system with novel relation annotation.
\newblock In \emph{Proceedings of the Thirteenth Language Resources and Evaluation Conference (LREC 2022)}.

\bibitem[{Chika et~al.(2024)Chika, Okahisa, Kodama, Huang, Murawaki, and Kurohashi}]{chika_2024_domain}
Taishi Chika, Taro Okahisa, Takashi Kodama, Yin~Jou Huang, Yugo Murawaki, and Sadao Kurohashi. 2024.
\newblock \href {https://aclanthology.org/2024.lrec-main.471/} {Domain transferable semantic frames for expert interview dialogues}.

\bibitem[{Ding et~al.(2023)Ding, Qin, Liu, Chia, Joty, Li, and Bing}]{ding2023gpt3}
Bosheng Ding, Chengwei Qin, Linlin Liu, Yew~Ken Chia, Shafiq Joty, Boyang Li, and Lidong Bing. 2023.
\newblock \href {https://arxiv.org/abs/2212.10450} {Is gpt-3 a good data annotator?}
\newblock \emph{Preprint}, arXiv:2212.10450.

\bibitem[{Ebner et~al.(2020)Ebner, Xia, Culkin, Rawlins, and Van~Durme}]{ebner-etal-2020-multi}
Seth Ebner, Patrick Xia, Ryan Culkin, Kyle Rawlins, and Benjamin Van~Durme. 2020.
\newblock \href {https://doi.org/10.18653/v1/2020.acl-main.718} {Multi-sentence argument linking}.
\newblock In \emph{Proceedings of the 58th Annual Meeting of the Association for Computational Linguistics}, pages 8057--8077, Online. Association for Computational Linguistics.

\bibitem[{Kalyanpur et~al.(2020)Kalyanpur, Biran, Breloff, Chu-Carroll, Diertani, Rambow, and Sammons}]{kalyanpur2020opendomainframesemanticparsing}
Aditya Kalyanpur, Or~Biran, Tom Breloff, Jennifer Chu-Carroll, Ariel Diertani, Owen Rambow, and Mark Sammons. 2020.
\newblock \href {https://arxiv.org/abs/2010.10998} {Open-domain frame semantic parsing using transformers}.
\newblock \emph{Preprint}, arXiv:2010.10998.

\bibitem[{Kingsbury and Palmer(2002)}]{kingsbury-palmer-2002-treebank}
Paul Kingsbury and Martha Palmer. 2002.
\newblock \href {http://www.lrec-conf.org/proceedings/lrec2002/pdf/283.pdf} {From {T}ree{B}ank to {P}rop{B}ank}.
\newblock In \emph{Proceedings of the Third International Conference on Language Resources and Evaluation ({LREC}{'}02)}, Las Palmas, Canary Islands - Spain. European Language Resources Association (ELRA).

\bibitem[{Ma et~al.(2023)Ma, Cao, Hong, and Sun}]{ma-etal-2023-large}
Yubo Ma, Yixin Cao, Yong Hong, and Aixin Sun. 2023.
\newblock \href {https://doi.org/10.18653/v1/2023.findings-emnlp.710} {Large language model is not a good few-shot information extractor, but a good reranker for hard samples!}
\newblock In \emph{Findings of the Association for Computational Linguistics: EMNLP 2023}, pages 10572--10601, Singapore. Association for Computational Linguistics.

\bibitem[{Okahisa et~al.(2022)Okahisa, Tanaka, Kodama, Huang, and Kurohashi}]{okahisa_2022_constructing}
Taro Okahisa, Ribeka Tanaka, Takashi Kodama, Yin~Jou Huang, and Sadao Kurohashi. 2022.
\newblock \href {https://aclanthology.org/2022.lrec-1.335} {Constructing a culinary interview dialogue corpus with video conferencing tool}.
\newblock In \emph{Proceedings of the Thirteenth Language Resources and Evaluation Conference}, pages 3131--3139, Marseille, France. European Language Resources Association.

\bibitem[{OpenAI(2024)}]{openai2024gpt4technicalreport}
OpenAI. 2024.
\newblock \href {https://arxiv.org/abs/2303.08774} {Gpt-4 technical report}.
\newblock \emph{Preprint}, arXiv:2303.08774.

\bibitem[{Skachkova and Kruijff-Korbayova(2021)}]{skachkova-kruijff-korbayova-2021-automatic}
Natalia Skachkova and Ivana Kruijff-Korbayova. 2021.
\newblock \href {https://aclanthology.org/2021.iwcs-1.10} {Automatic assignment of semantic frames in disaster response team communication dialogues}.
\newblock In \emph{Proceedings of the 14th International Conference on Computational Semantics (IWCS)}, pages 93--109, Groningen, The Netherlands (online). Association for Computational Linguistics.

\bibitem[{Sun et~al.(2023)Sun, Dong, Li, Wan, Wang, Zhang, Li, Cheng, Lyu, Wu, and Wang}]{sun2023pushinglimitschatgptnlp}
Xiaofei Sun, Linfeng Dong, Xiaoya Li, Zhen Wan, Shuhe Wang, Tianwei Zhang, Jiwei Li, Fei Cheng, Lingjuan Lyu, Fei Wu, and Guoyin Wang. 2023.
\newblock \href {https://arxiv.org/abs/2306.09719} {Pushing the limits of chatgpt on nlp tasks}.
\newblock \emph{Preprint}, arXiv:2306.09719.

\bibitem[{Vaswani et~al.(2023)Vaswani, Shazeer, Parmar, Uszkoreit, Jones, Gomez, Kaiser, and Polosukhin}]{vaswani2023attentionneed}
Ashish Vaswani, Noam Shazeer, Niki Parmar, Jakob Uszkoreit, Llion Jones, Aidan~N. Gomez, Lukasz Kaiser, and Illia Polosukhin. 2023.
\newblock \href {https://arxiv.org/abs/1706.03762} {Attention is all you need}.
\newblock \emph{Preprint}, arXiv:1706.03762.

\bibitem[{Wan et~al.(2023)Wan, Cheng, Mao, Liu, Song, Li, and Kurohashi}]{wan2023gptre}
Zhen Wan, Fei Cheng, Zhuoyuan Mao, Qianying Liu, Haiyue Song, Jiwei Li, and Sadao Kurohashi. 2023.
\newblock Gpt-re: In-context learning for relation extraction using large language models.
\newblock In \emph{Proceedings of the 2023 Conference on Empirical Methods in Natural Language Processing}, pages 3534--3547.

\bibitem[{Wang et~al.(2023{\natexlab{a}})Wang, Sun, Li, Ouyang, Wu, Zhang, Li, and Wang}]{wang2023gptner}
Shuhe Wang, Xiaofei Sun, Xiaoya Li, Rongbin Ouyang, Fei Wu, Tianwei Zhang, Jiwei Li, and Guoyin Wang. 2023{\natexlab{a}}.
\newblock Gpt-ner: Named entity recognition via large language models.
\newblock \emph{arXiv preprint arXiv:2304.10428}.

\bibitem[{Wang et~al.(2021)Wang, Liu, Xu, Zhu, and Zeng}]{wang-etal-2021-want-reduce}
Shuohang Wang, Yang Liu, Yichong Xu, Chenguang Zhu, and Michael Zeng. 2021.
\newblock \href {https://doi.org/10.18653/v1/2021.findings-emnlp.354} {Want to reduce labeling cost? {GPT}-3 can help}.
\newblock In \emph{Findings of the Association for Computational Linguistics: EMNLP 2021}, pages 4195--4205, Punta Cana, Dominican Republic. Association for Computational Linguistics.

\bibitem[{Wang et~al.(2023{\natexlab{b}})Wang, Kordi, Mishra, Liu, Smith, Khashabi, and Hajishirzi}]{wang_2023_selfinstruct}
Yizhong Wang, Yeganeh Kordi, Swaroop Mishra, Alisa Liu, Noah~A. Smith, Daniel Khashabi, and Hannaneh Hajishirzi. 2023{\natexlab{b}}.
\newblock \href {https://doi.org/10.18653/v1/2023.acl-long.754} {Self-instruct: Aligning language models with self-generated instructions}.
\newblock In \emph{Proceedings of the 61st Annual Meeting of the Association for Computational Linguistics (Volume 1: Long Papers)}, pages 13484--13508, Toronto, Canada. Association for Computational Linguistics.

\bibitem[{Zhang et~al.(2023)Zhang, Guti{\'e}rrez, and Su}]{zhang2023aligning}
Kai Zhang, Bernal~Jim{\'e}nez Guti{\'e}rrez, and Yu~Su. 2023.
\newblock Aligning instruction tasks unlocks large language models as zero-shot relation extractors.
\newblock \emph{arXiv preprint arXiv:2305.11159}.

\end{thebibliography}

\appendix

\section{Appendix}
\label{sec:appendix}

\subsection{Prompt For Pseudo-dialogue Generation By LLM} 
\label{sec:pd_prompt_appendix}
\begin{figure}[h]
  \includegraphics[width=\columnwidth]{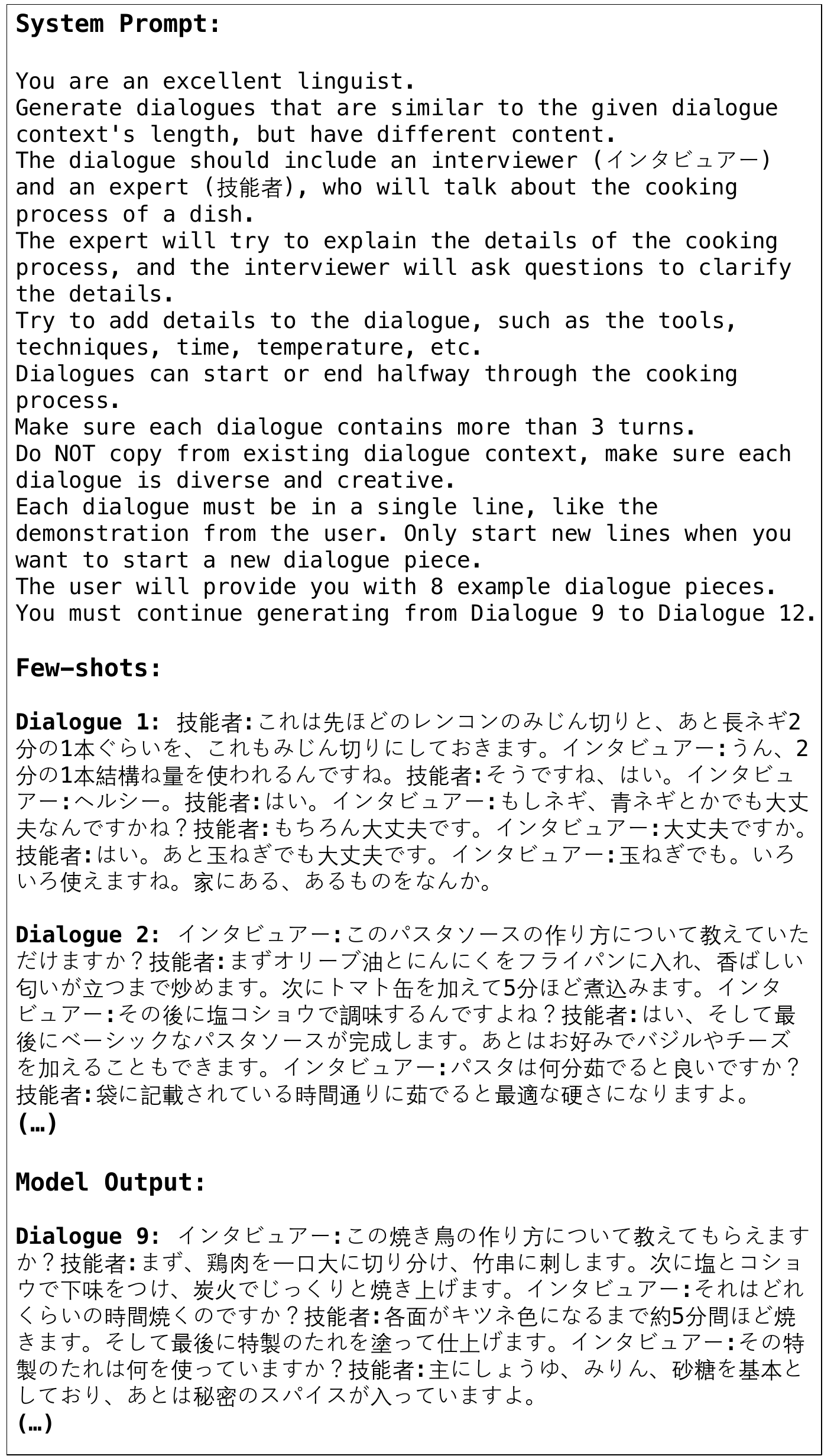}
  \caption {The prompt design for pseudo-dialogue generation. In this example, Dialogue 1 is a human dialogue, and Dialogue 2 is a previously generated pseudo-dialogue. }
  \label{fig:pd_prompt}
\end{figure}

An example prompt for pseudo-dialogue generation is shown in Figure~\ref{fig:pd_prompt}.

\subsection{Prompt For LLM SFA Labeling} 
\label{sec:sfa_prompt_appendix}

\begin{figure}[h]
  \includegraphics[width=\columnwidth]{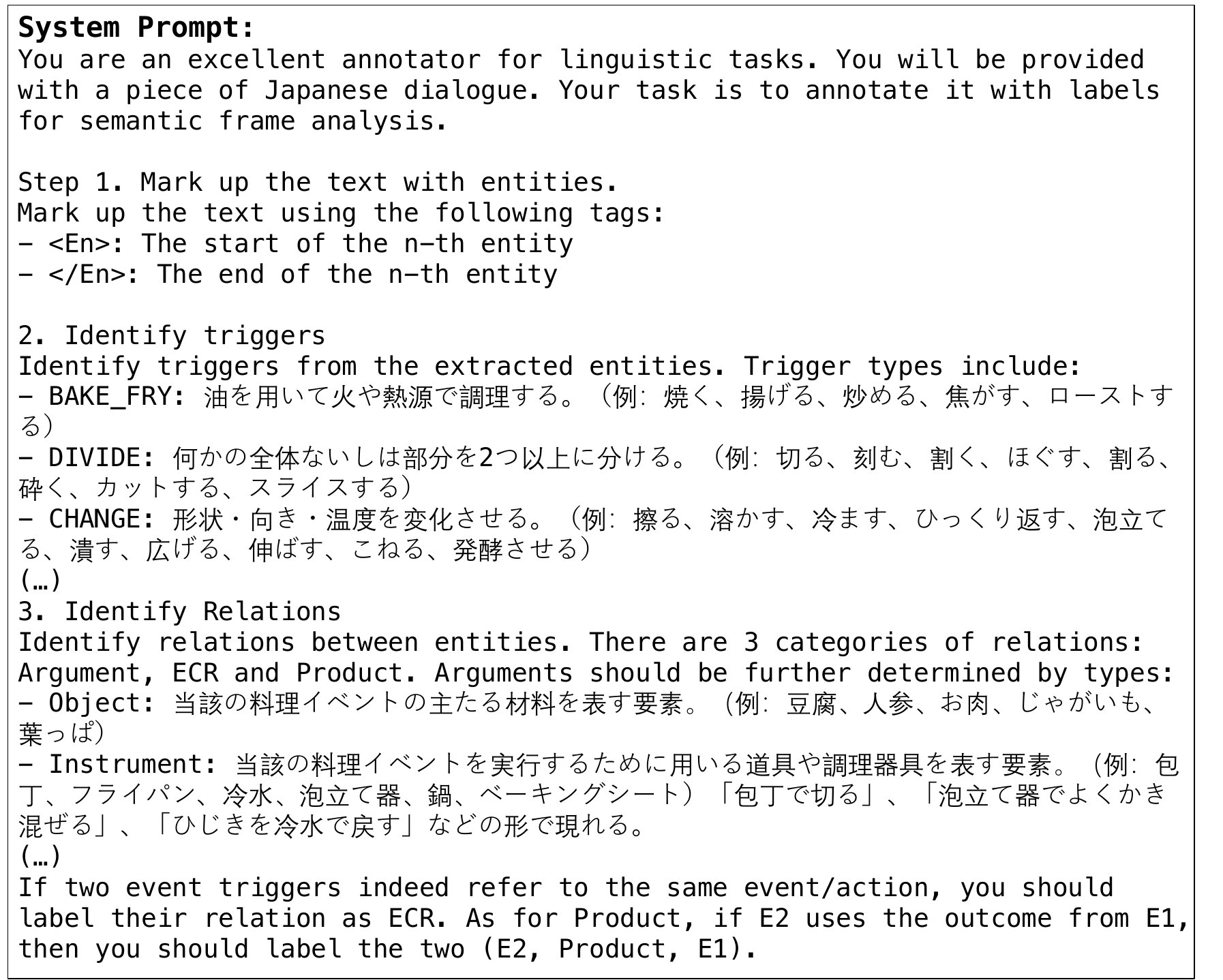}
  \caption {The system prompt for the LLM. Some type definitions are omitted. }
  \label{fig:prompt1}
\end{figure}

\begin{figure}[h]
  \includegraphics[width=\columnwidth]{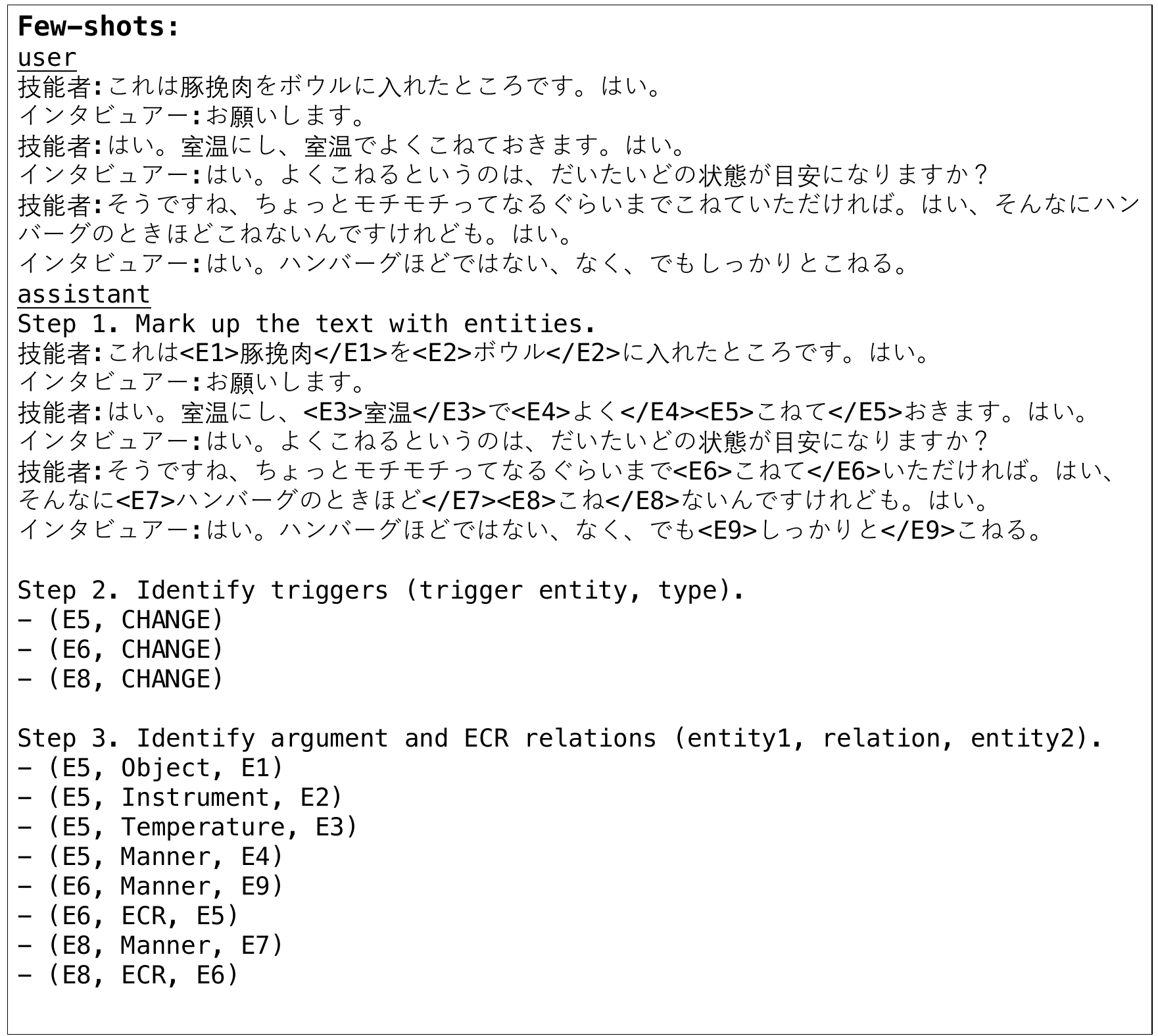}
  \caption {The few-shot demonstration for the LLM. We show one example here. }
  \label{fig:prompt2}
\end{figure}

\begin{figure}[h]
  \includegraphics[width=\columnwidth]{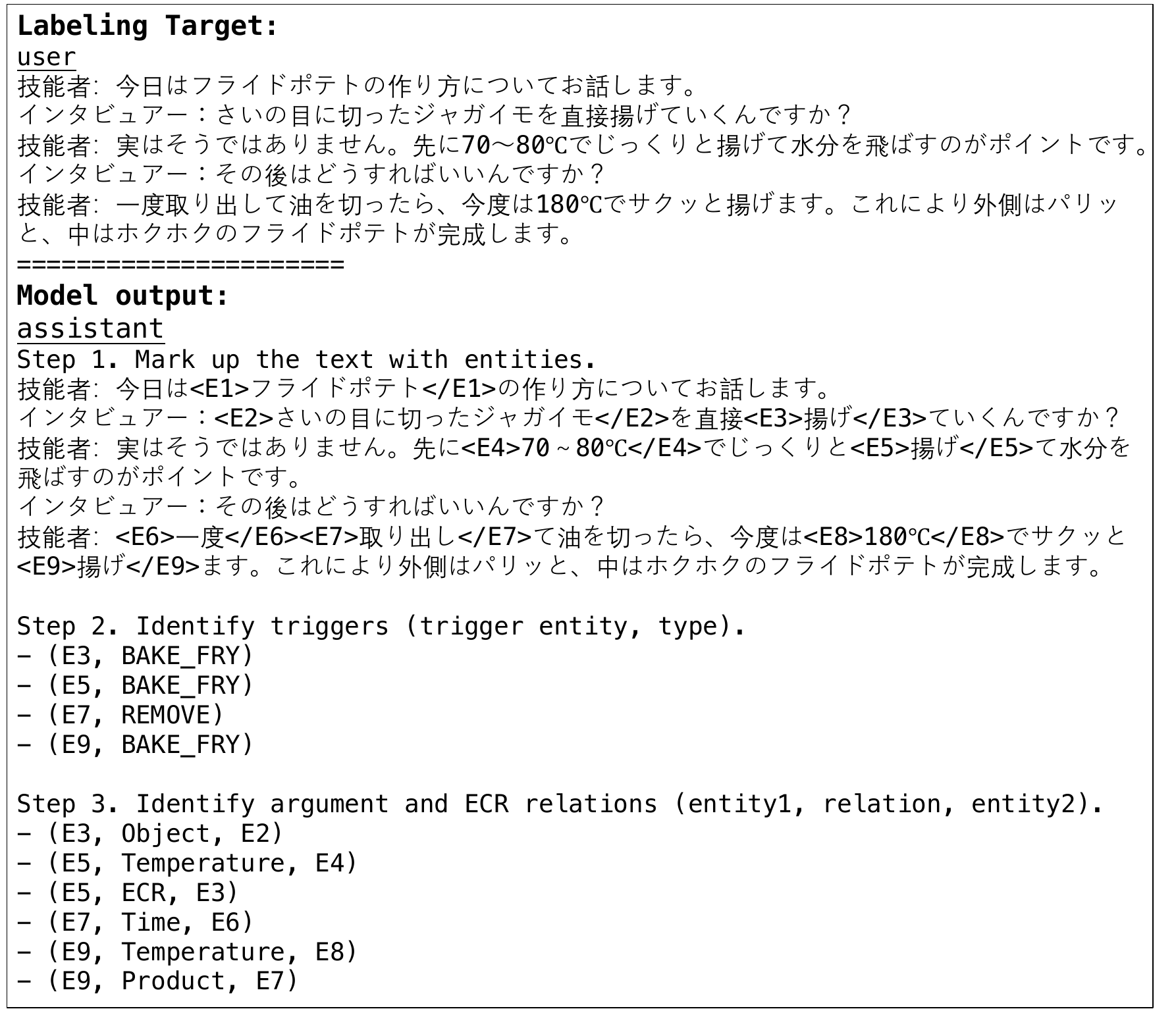}
  \caption {The input labeling target and an actual labeling output from the LLM. This is an example from the Pseudo-Pseudo data.}
  \label{fig:prompt3}
\end{figure}

The prompt provided to the LLM for SFA labeling is shown in Figure~\ref{fig:prompt1},~\ref{fig:prompt2},~\ref{fig:prompt3}.

\subsection{Length Distribution of Pseudo-dialogues} \label{sec:length_dist}
\begin{figure}[h]
  \includegraphics[width=\columnwidth]{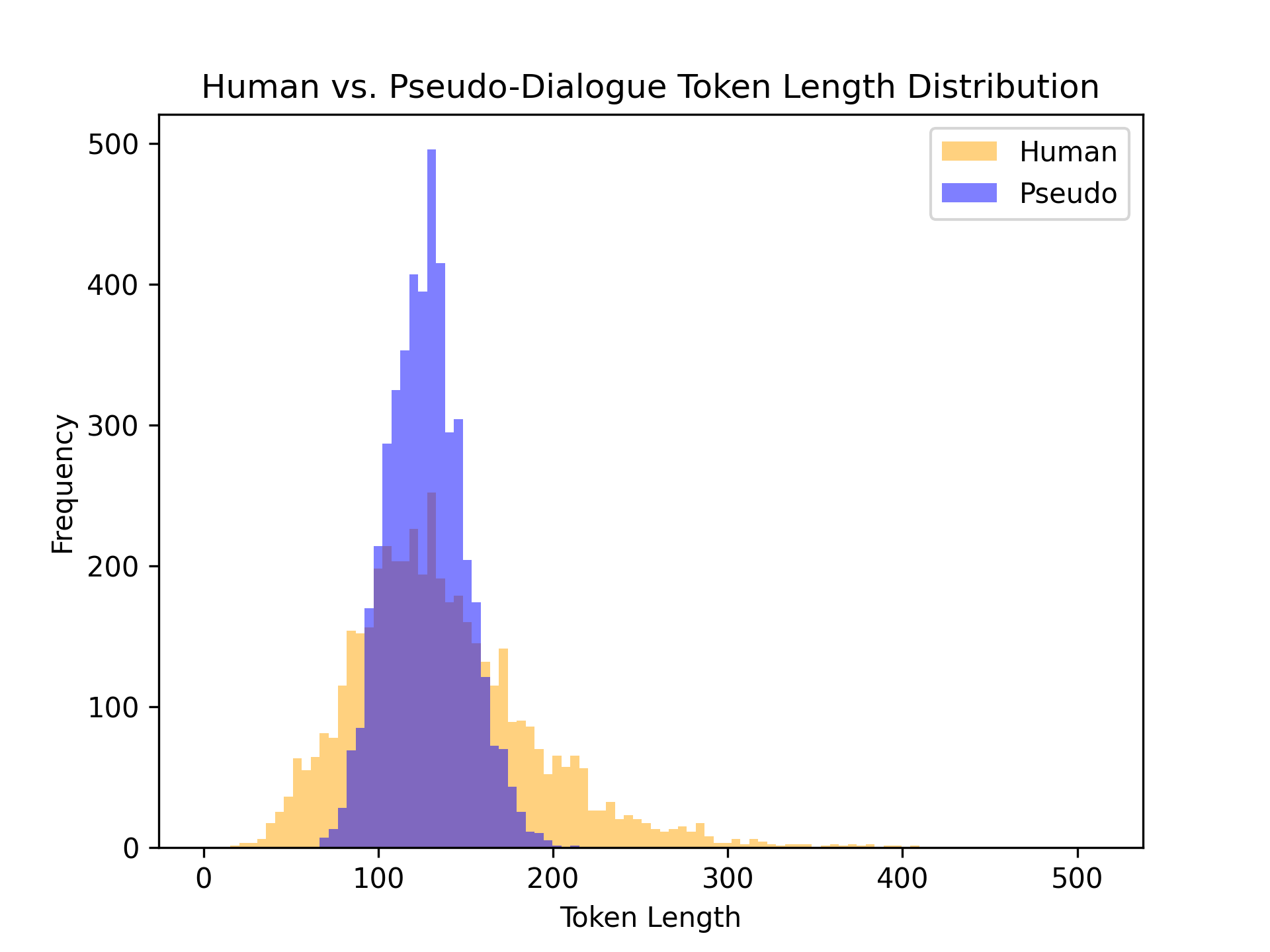}
  \caption {The length distributions of human and pseudo-dialogues.}
  \label{fig:length_dist}
\end{figure}

We present the length distributions of human dialogues and pseudo-dialogues. We observed that GPT-4 generally followed the length specification in the instruction, resulting in an average length of 127 tokens (token count by Japanese DeBERTa-V2 tokenizer) compared to an average of 136 tokens in human dialogue sessions. Moreover, pseudo-dialogues have a more compact distribution, which means there are fewer extremely short or long outliers.

\subsection{Label Distribution in Pseudo-dialogues} \label{sec:label_dist}

\begin{figure*}[t]
  \centering
  \includegraphics[width=0.8\linewidth]{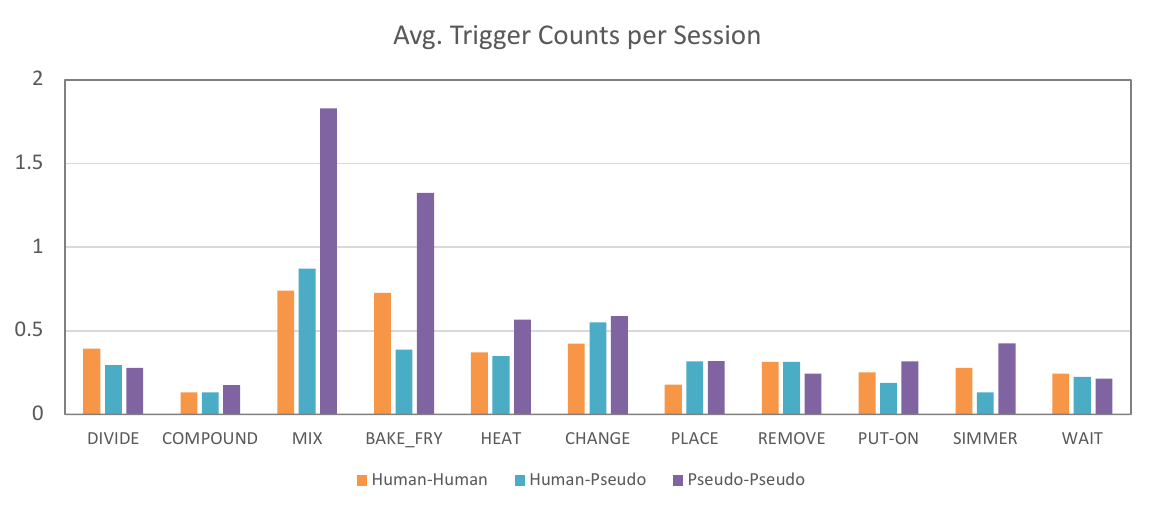} 
  % \hfill
  \includegraphics[width=0.57\linewidth]{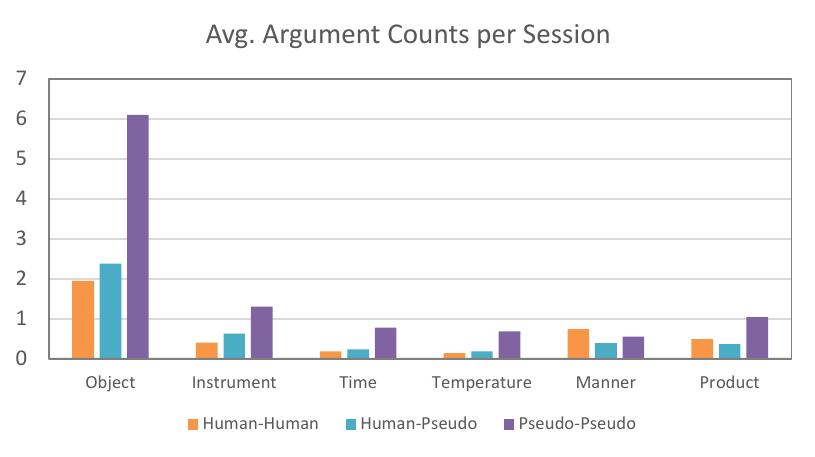}
  \caption {Trigger and argument label distribution in Human-Human, Human-Pseudo and Pseudo-Pseudo data.}
  \label{fig:frame_dist}
\end{figure*}

We present the label distributions across three data types: Human-Human, Human-Pseudo, and Pseudo-Pseudo in Figure~\ref{fig:frame_dist}. When comparing Human-Human to Human-Pseudo, we observe that replacing human labelers with GPT-4 leads to fluctuations in certain label types. Specifically, there is a decrease in types such as "BAKE\_FRY" and "SIMMER" in triggers and "Manner" in arguments, and an increase in types like "PLACE" in triggers and "Instrument" in arguments. While we believe that these fluctuations will not be a significant issue, it is important to point out that in addition to the fluctuations, the labels generated by GPT-4 may not be accurate either.

When comparing Human-Pseudo to Pseudo-Pseudo, we observe that replacing human dialogues with pseudo-dialogues leads to a higher frequency of certain types than in human dialogues. For example, types like "MIX" and "BAKE\_FRY" in triggers and all argument types appear more frequently. This increase occurs because GPT-4 tends to fit a whole story into a pseudo-dialogue, resulting in a higher overall entity count.  In contrast, human dialogues are heuristically cut into smaller sessions, which can lead to fewer entities per session. Also, the increase in trigger types "MIX" and "BAKE\_FRY" indicates that GPT-4 tends to mention these specific events, creating a bias toward specific topics.

\subsection{Training Hyperparameters for the SLM}
\label{sec:hyperparams}
We adopted JaMIE \cite{cheng-etal-2022-jamie} as our SLM for SFA. For the encoder, we used a pre-trained Japanese DeBERTa-V2-base model with an encoder learning rate of 2e-5 and a relation decoder learning rate of 1e-2, without a learning rate schedule.\footnote{\href{https://huggingface.co/ku-nlp/deberta-v2-base-japanese}{https://huggingface.co/ku-nlp/deberta-v2-base-japanese}} The model was trained for up to 30 epochs, and the best checkpoint was selected based on the highest validation weighted F1 score. The validation and test sets are defined in the EIDC dataset with sizes of 269 and 379 dialogue sessions, respectively.

\subsection{Discussion: Few-shot LLM or training an SLM?}
The few-shot SFA labeling performance of GPT-4 is depicted in Figures~\ref{fig:rp} and~\ref{fig:pp}, shown in black dotted lines. The result shows that with just \$200 of Pseudo-Pseudo data, we can train a much smaller SLM that matches GPT-4's performance on SFA. Therefore, it is more advisable to train an SLM using purely synthetic data, i.e. the Pseudo-Pseudo data, avoiding the running cost and stability issue of an LLM (Section~\ref{llm-instability}). 

\end{document}